\documentclass[pdflatex, sn-mathphys-num, iicol]{sn-jnl}

\usepackage{graphicx}%
\usepackage{multirow}%
\usepackage{amsmath,amssymb,amsfonts}%
\usepackage{amsthm}%
\usepackage[title]{appendix}%
\usepackage{xcolor}%
\usepackage{textcomp}%
\usepackage{manyfoot}%
\usepackage{booktabs}%
\usepackage[ruled,vlined]{algorithm2e}
\usepackage{algpseudocode}%
\usepackage{listings}%
\usepackage{subcaption}
\usepackage{xfrac}
\usepackage{nicefrac}
\usepackage{comment}
\usepackage{cuted}
\usepackage{multicol}
\usepackage[belowskip=0pt,aboveskip=5pt]{caption}

\theoremstyle{thmstyleone}%
\theoremstyle{thmstyletwo}%

\theoremstyle{thmstylethree}%

\begin{document}

\title[Article Title]{Koopman-Based Nonlinear Identification and Model Predictive Control of a Turbofan Engine}

\author*[1]{\fnm{David} \sur{Grasev}}\email{david.grasev@unob.cz}

\affil*[1]{\orgdiv{Department of Aviation Technology}, \orgname{University of Defence}, \orgaddress{\city{Brno}, \state{Czech Republic}}}

\abstract{This paper investigates Koopman operator-based approaches for multivariable control of a two-spool turbofan engine. A physics-based component-level model is developed to generate training data and validate the controllers. A meta-heuristic extended dynamic mode decomposition is adapted, with a cost function designed to accurately capture both spool-speed dynamics and the engine pressure ratio (EPR), enabling the construction of a single Koopman model that can be reused across multiple control strategies. Using the identified time-varying Koopman model, an adaptive Koopman-based model predictive controller (AKMPC) with a disturbance observer is developed and compared with a Koopman-based feedback linearization controller (K-FBLC) and its integrator-augmented version (K-FBLC-I). The Koopman representation further enables nonlinear GTE output limiters, such as rotor-acceleration and turbine-inlet-temperature limits, to be expressed as linear constraints in the AKMPC. The controllers are evaluated for two control configurations of spool speeds and EPR, under both sea-level and varying flight conditions. The results demonstrate that the proposed identification approach enables accurate predictions of both spool speeds and EPR, allowing the Koopman model to be reused flexibly across different control formulations. While all strategies achieve comparable performance in sea-level conditions, the AKMPC demonstrates improved performance under varying flight conditions due to its ability to capture nonlinear dynamics, handle constraints, and compensate for model mismatch. Moreover, the EPR control strategy improves the thrust response. The study highlights the applicability of Koopman-based control and the advantages of the AKMPC framework for robust turbofan engine control.}

\keywords{Data-driven modeling, Feedback linearization, Gas turbine engines, Koopman operator, Koopman eigenfunctions, Nonlinear dynamics, Nonlinear control, Model predictive control}

\maketitle

\section{Introduction} \label{sec1}
Modern aviation relies heavily on turbofan gas turbine engines (GTEs), which provide efficient and reliable propulsion across a wide range of operating conditions. With a growing emphasis on fuel economy, emission reduction, and operational flexibility, the performance requirements for these engines have become increasingly demanding. Meeting these requirements necessitates advanced control methodologies capable of handling complex nonlinear dynamics and disturbances. The main control challenges arise from strict performance and safety requirements, including rapid response to setpoint changes while maintaining safe operation \cite{MattiglyBook, Jaw2009, Garg2013, Lv2022-GTE-Rev}.

Typically, only one spool speed is controlled using fuel flow, while other quantities, such as the second spool speed, gas temperature, and surge margin, are maintained within their limits by a protection system \cite{Jaw2009, Garg2013}. These systems feature devices such as variable vanes and bleed valves that regulate rotor-blade inflow, thereby introducing a multivariable control problem \cite{MattiglyBook, Jaw2009, Garg2013, Cheng2021-Hinf}. However, fuel flow can be combined with additional control variables to improve the GTE dynamic response, particularly its response speed and fuel consumption. Examples include variable stator vanes \cite{Cheng2021-Hinf} and nozzle area \cite{Garg1989, Garg2013}. Controlled outputs include the low-pressure spool speed $N_1$, the high-pressure spool speed $N_2$, the engine pressure ratio (EPR) $\Pi_\mathrm{EPR}$, or thrust $F$, which can enter the control problem either indirectly through spool speed or EPR allocation, or directly \cite{Wei2025-DTC, Zhu2022-DTC-LPV}. 

Strong aerodynamic coupling between engine outputs poses a major challenge for classical proportional-integral (PI) control approaches, which try to address it by employing a decoupling system derived from the inverse dynamics of the controller-engine connection and by compensating for rotor cross-talk \cite{Li2025-RBF-LSTM-ANN-Hinf}. However, the decoupler design becomes significantly more complex when gain scheduling and nonlinear methods are considered. Linear quadratic regulators (LQRs) can compensate for coupling while providing optimal controller tuning, especially in a robust gain-scheduling setup. 

Many advanced nonlinear control approaches build on component-level models (CLMs). CLMs allow linearization around multiple operating points, e.g., via a small perturbation method, to obtain linearized models \cite{Pang2020-CLM, Pang2020-CLM2}. On-board CLMs enable, e.g., optimization of protection limiters, improvement of transient response, and balancing of safety and performance \cite{Pang2021-Limits}. In \cite{Yao2025-DualLoop}, a performance-seeking controller with a dual-loop structure was introduced. The inner loop stabilized the system, while the outer loop uses a CLM to optimize the GTE’s variable geometry, minimizing fuel consumption and addressing the multi-objective nature of GTE control. However, CLMs are computationally expensive for real-time applications. Additionally, obtaining or matching component characteristics can be cumbersome, especially for multi-rotor configurations, because measured data or geometry for numerical simulation are often proprietary \cite{GrasevSpringer2026}.

Robust gain-scheduling approaches using linear parameter-varying (LPV) models were investigated in the literature. Cheng et al. proposed a robust $\mathcal{H}_\infty$ gain-scheduled controller designed using a polynomial LPV model \cite{Cheng2021-Hinf}. Zhu et al. designed a robust direct thrust controller based on an affine LPV model and linear matrix inequalities \cite{Zhu2022-DTC-LPV}. The model assumed fuel flow control in two regimes at the minimum and maximum nozzle areas. An on-board adaptive polytopic LPV model was developed in \cite{Liu2023-LPV}, which features an extended Kalman filter (EKF) to improve performance. Subsequently, a gain-scheduled robust controller was designed using this model. In \cite{Chen2023-Adaptive} and \cite{Chen2023-Onboard}, an adaptive on-board model with an unscented Kalman filter was developed to address the estimation divergence issue of EKF-based models, and a self-adapting performance recovery control was proposed. The models also enable the estimation of health parameters that capture engine degradation. Alternatives to LPV models include, e.g., Hybrid Wiener models \cite{Wei2021-Wiener, Wei2022-Wiener-ATP}, or off-equilibrium linearization \cite{Yang2024-OffEqLin}.

An ongoing increase in accessible on-board computational power enables the employment of more advanced nonlinear control methods, such as dynamic inversion \cite{Singh2022-Robust-DynInversion}, sliding-mode control \cite{Palmieri2021-SMC}, feedback linearization \cite{Bonfiglio2017, Erario2020-FBL-SINDy}, or model predictive control (MPC), which utilizes online optimization and a model of the GTE to find a sequence of control inputs, minimizing a specified cost function across a predictive horizon. MPC shows a strong potential for multivariable control, making it highly suitable for turbofan GTEs. In \cite{Brunell2004}, MPC was proposed for turbofan engine control, with an online adaptation of the linearized model. Montazeri et al. proposed MPC with feedback corrections for the control of turbofan engines, and performed hardware-in-the-loop simulations, discussing real-time capabilities \cite{Montazeri2019-MPC}. In \cite{Song2025-NMPC}, the authors propose a long short-term memory model with support vector regression to facilitate multivariable nonlinear MPC control of a helicopter turboshaft engine. A nonlinear MPC framework was combined with a deep neural network model in \cite{Zheng2019-DNN-MPC} for direct thrust control. Further MPC applications to GTEs are found, e.g., in \cite{Pang2021-DTC-MPC, Pang2021-MPC, Ibrahem2021-NARX-MPC, Wang2022-ANN-MPC-fuzzy, Ji2022-MPC}. 

Some of the above-mentioned papers feature machine learning-based black-box models, such as neural networks, which suffer from low interpretability. The sparse identification of nonlinear dynamics (SINDy) provides more interpretable models and was employed in \cite{Erario2020-FBL-SINDy, Momin2022-SINDy-ThrustEst} to obtain an interpretable nonlinear control-affine model of a small turbojet engine, enabling design of a feedback linearization controller. 

Many of the aforementioned approaches rely on local linearization, extensive scheduling, or black-box models, which complicates unified multivariable control design and limits model reuse across different control formulations. LPV models' ability to capture the highly nonlinear dynamics of turbofan GTEs, especially near their operating limits, is limited. Furthermore, solving nonlinear MPC problems is often very expensive. This motivates the search for modeling frameworks that can capture nonlinear dynamics while retaining a structure suitable for simpler control systems. 

In recent years, the Koopman operator has emerged as a tractable tool that enables global linearization of nonlinear dynamics via a nonlinear coordinate transformation (lifting) of the state space to a new observable space \cite{Mauroy2020}. The linearization can hold in the entire basin of attraction or in subsets of the system well covered by the data. In the case of turbofan GTEs, the subset is represented by the region of safe operation bounded by the surge/stall, temperature, spool speed, and combustion limits. Moreover, a linear Kalman filter can be designed in the lifted state space using simple Riccati-based synthesis, as opposed to nonlinear EKF or unscented filters \cite{Surana2016-KoopmanKalman, Surana2020}.

A key advantage of the Koopman framework lies in its linear representation in the observable space, enabling the utilization of linear optimal control techniques, such as LQR or MPC \cite{Mauroy2020, Proctor2018}. The parameter-varying Koopman eigenfunction models and LQR control of a variety of nonlinear systems were discussed, e.g., in \cite{Kaiser2021}. The results demonstrated that LPV Koopman-based controllers outperform linear LQR and feedback linearization controllers. In our previous work \cite{GrasevSpringer2026}, SINDy was employed to derive a control-affine model of a single-spool turbojet GTE with a subsequent transformation to a Koopman model via temporal identification of Koopman eigenvalues and eigenfunctions. The LQR controller with integral action outperformed gain-scheduled PI and internal-model controllers, highlighting the applicability of Koopman-based control for GTEs. Koopman MPC (KMPC) was first proposed in \cite{Korda2018-KMPC}, which outlined its strengths and provided theoretical analysis. The KMPC has been shown to outperform local and Carleman linearization-based MPC on examples of the van der Pol oscillator, a bilinear motor, and a shallow-water partial differential equation. In \cite{Korda2020}, the authors proposed an alternative Koopman eigenfunction identification method and showcased KMPC for the Duffing oscillator. 

Numerous extensions and modifications of (K)MPC have been proposed. One approach to compensate for disturbances and model mismatch is offset-free MPC, which was introduced in \cite{Pannocchia2002-OF-MPC, Pannocchia2015-OF-MPC-New} and employs a Kalman-filter-based disturbance observer (DO). This idea has also been incorporated into Koopman-based MPC. In \cite{Chen2022-OffsetFree-KMPC, Li2025-OF-KMPC-Robots}, offset-free KMPC was applied to the control of soft
robots. The authors compared a basic KMPC with a KMPC equipped with a Kalman-based DO across several tasks, and the results showed that the disturbance observer improved the closed-loop
performance. Further recent works on offset-free and disturbance-rejection Koopman MPC include, e.g., \cite{Pan2024-OF-KMPC-AirSystem, Schimperna2025}.

While the DO-based approaches typically employ fixed time-invariant Koopman models, model mismatch can also be addressed via adaptive Koopman methods, where the Koopman model is updated online. In \cite{Wu2026-Adaptive-ANN-KMPC}, an autoencoder-based KMPC with a recursive least-squares adaptation strategy was employed for the control of distributed-drive electric vehicles. At each control instance, the updated model was held constant across the prediction horizon, keeping the linear MPC structure. The results showed that the adaptive KMPC outperformed several alternative methods, including nonlinear MPC and Gaussian-process-regression MPC. Other works on adaptive KMPC include, e.g., \cite{Dittmer2022-Adaptive-KMPC, Singh2025-AdaptiveKMPC}, while related robust Koopman MPC variants include, e.g., stochastic KMPC \cite{Kim2025-SKMPC} and tube KMPC \cite{Zhou2025-Dual-KMPC}.

Although the Koopman‑based predictive control has been investigated for various nonlinear systems, its application to multivariable constrained control of turbofan engines remains limited in the open literature. The present paper addresses this gap by developing an adaptive KMPC (AKMPC) framework for a two-spool turbofan engine. The proposed approach uses a time-varying low-order Koopman system specifically designed to accurately predict spool speeds and the EPR. The model is identified offline using the meta-heuristic extended dynamic mode decomposition (MH-EDMD) method from \cite{GrasevAccess2025}. In contrast to methods that update the Koopman operator or dictionary online, the proposed method keeps the identified observable functions fixed. The KMPC online adaptation comes from evaluating the state-dependent matrices of the identified Koopman model at the current operating point. The resulting time-varying linear model is then held constant over the prediction horizon, which preserves the quadratic-programming formulation, while allowing the prediction model to vary with the engine operating condition. In addition, a Kalman filter-based DO is included to compensate for mismatch caused by variations in flight conditions and modeling uncertainty. 

The present work extends the authors' previous Koopman-based GTE study in \cite{GrasevSpringer2026}, which considered a single-spool turbojet engine, to a two-spool turbofan engine. This extension introduces strong rotor coupling, two manipulated variables, and thus a coupled multivariable constrained control problem, motivating the use of MPC. Compared with \cite{GrasevAccess2025}, where the MH-EDMD identification framework was introduced, the present paper modifies the identification objective for multi-output turbofan prediction and embeds the resulting Koopman model in a constrained AKMPC framework with actuator dynamics and disturbance-observer compensation. The proposed framework further demonstrates that traditional nonlinear GTE output limiters can be handled directly through linear inequality constraints thanks to the Koopman framework.

The main contributions of this paper are summarized as follows:
\begin{itemize}
    \item Development of a time-varying Koopman model of a turbofan GTE specifically tailored to the accurate prediction of both spool speeds and EPR, using MH-EDMD, with a modified multi-criteria objective function.

    \item Application of AKMPC with a disturbance observer and linear constraints, representing the nonlinear limiters, to control of a turbofan GTE, and its comparative evaluation using a proposed Koopman-based feedback linearization controller (K-FBLC) and its integrator-augmented version (K-FBLC-I) as benchmarks within a unified modeling framework.
    
    \item Demonstration that a single identified Koopman model can be consistently reused across multiple control strategies without structural modifications, including $N_1-N_2$ and $\Pi_\mathrm{EPR}-N_1$ control, showcasing its flexibility for multi-output control design.
    
    \item Comprehensive evaluation under both sea-level conditions and varying flight conditions, demonstrating accurate tracking of spool speeds and EPR, and showing robustness of the AKMPC in the presence of operating condition variations. 
\end{itemize}

The rest of the paper is organized as follows: Section \ref{sec2} introduces an in-house turbofan GTE physics-based model. In Section \ref{sec3}, fundamentals of the Koopman operator theory are described, and Section \ref{sec4} provides an overview of the MH-EDMD and the corresponding cost function. Section \ref{sec5} is devoted to the description of AKMPC, K-FBLC, and K-FBLC-I controllers. The simulation results are summarized and discussed in Section \ref{sec6}, and the main conclusions are presented in Section \ref{sec7}.

\subsection*{Notation}
The main notation is summarized in Table \ref{tab_notation}.
\begin{table*}[!ht]
\centering
\caption{Summary of the main notation used in the manuscript.}
\label{tab_notation}
\begin{tabular}{lll}
\toprule
Symbol & Dimension / unit & Description \\
\midrule
$N_1,N_2$ & RPM & Low- and high-pressure spool speeds \\
$W_\mathrm{f}$ & kg/s & Fuel flow \\
$A_\mathrm{n}$ & m$^2$ & Nozzle area \\
$\Pi_\mathrm{EPR}$ & -- & Engine pressure ratio \\
$F$ & N & Net thrust \\
$\mathbf{x}$ & $\mathbb{R}^{2}$ & State vector, $\mathbf{x}=[N_1,N_2]^\top$ \\
$\mathbf{u}$ & $\mathbb{R}^{2}$ & Input vector, $\mathbf{u}=[W_\mathrm{f},A_\mathrm{n}]^\top$ \\
$\mathbf{y}$ & $\mathbb{R}^{n_y}$ & Output vector \\
$\mathbf{\Psi(x)}$ & $\mathbb{R}^{n_\Psi}$ & Observable functions used in EDMD \\
$\mathbf{\Phi(x)}$ & $\mathbb{R}^{n_\Phi}$ & Koopman eigenfunctions \\
$\mathbf{A}$ & $\mathbb{R}^{n_\Psi\times n_\Psi}$ & Continuous-time Koopman system matrix \\
$\mathbf{\Lambda}$ & $\mathbb{R}^{n_\Phi\times n_\Phi}$ & Koopman eigenvalue matrix \\
$\mathbf{G(x)}$ & $\mathbb{R}^{n_\Phi\times n_u}$ & State-dependent input matrix in eigenfunction coordinates \\
$\mathbf{C_\Phi}$ & $\mathbb{R}^{n_y\times n_\Phi}$ & Output reconstruction matrix in eigenfunction coordinates \\
$\mathbf{x}_\mathrm{act}$ & $\mathbb{R}^{2}$ & Actuator state vector \\
$\mathbf{z}$ & $\mathbb{R}^{n_\Phi+2}$ & Augmented Koopman-actuator state vector \\
$\mathbf{x_a}$ & $\mathbb{R}^{n_\Phi+n_y+2}$ & Output-augmented state vector for AKMPC \\
$\mathbf{\hat{d}}$ & $\mathbb{R}^{n_y}$ & Estimated output disturbance \\
$\mathbf{L}$ & -- & Kalman/observer gain matrix \\
$n_\mathrm{p},n_\mathrm{c}$ & -- & Prediction and control horizons \\
$\mathbf{Q_y,Q_T,R}$ & -- & Output-error, terminal, and input-increment MPC weighting matrices \\
\bottomrule
\end{tabular}
\end{table*}

\section{Turbofan GTE Component-Level Model} \label{sec2}
\begin{figure}[!ht]
    \centering
    \includegraphics[width=1\linewidth]{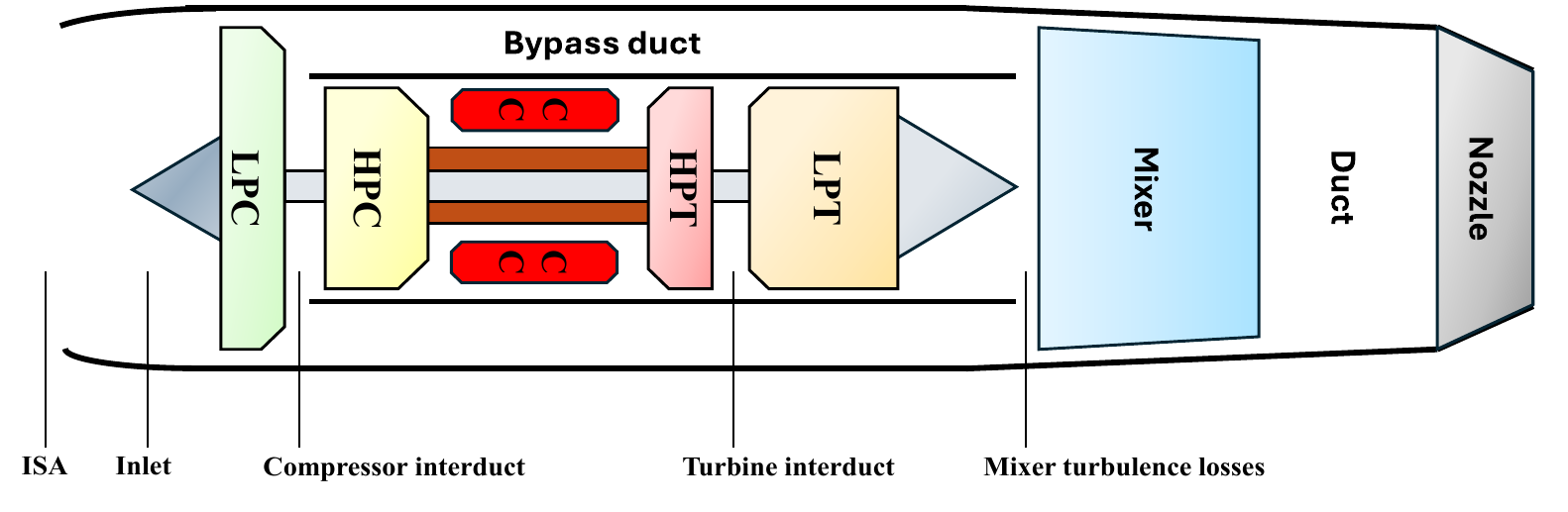}
    \caption{Turbofan GTE diagram with the main components considered in this paper.}
    \label{fig_GTE_diagram}
\end{figure}

To ensure full control over the model and controller implementation, an in-house thermodynamic CLM of a turbofan GTE was developed in MATLAB. The model is utilized to generate training data and evaluate control strategies. Therefore, it must be representative of the real GTE dynamics. A schematic diagram of the mixed-flow low-bypass turbofan engine is in Fig. \ref{fig_GTE_diagram}. The input variables are the fuel flow $W_\mathrm{f}$ and the nozzle area $A_\mathrm{n}$, the dynamic state variables are low-pressure and high-pressure spool speeds, $N_1$ and $N_2$, and the main output variables are the spool speeds, thrust $F$, and engine pressure ratio $\Pi_\mathrm{EPR}$.

\subsection{Component Models}
The effects of atmospheric and flight conditions were modeled using the International Standard Atmosphere and adiabatic relations between the inlet velocity and the static and total pressures and temperatures. 

The total pressure losses due to friction in the inlet, compressor interduct, splitter, bypass duct, combustion chamber, turbine interduct, mixer, and exhaust duct were modeled by their respective loss coefficients \cite{MattiglyBook}.

The component characteristics of compressors and turbines were extracted from the mixed-flow turbofan example from the GasTurb 15 simulation software \cite{GasTurb} and scaled to the design point of the target GTE. 

The parameters from low-pressure compressor (LPC) and high-pressure compressor (HPC) maps are obtained using $\beta$ interpolation presented in \cite{Grasev2024, Kurzke1996}.
\begin{equation} 
    \begin{aligned}
    \begin{bmatrix}
        W_\mathrm{LPC,c}, & \Pi_\mathrm{LPC}, & \eta_\mathrm{LPC}
    \end{bmatrix} &= f(N_\mathrm{1,c,LPC}, \beta_\mathrm{LPC})\,, \\
    \begin{bmatrix}
        W_\mathrm{HPC,c}, & \Pi_\mathrm{HPC}, & \eta_\mathrm{HPC}
    \end{bmatrix} &= f(N_\mathrm{2,c,HPC}, \beta_\mathrm{HPC})\,,
    \end{aligned}
\end{equation}
where $W_c$, $\Pi$, and $\eta$ denote the corrected mass flow, pressure ratio, and efficiency of the compressors, $\beta$ is an auxiliary interpolation variable, and $N_\mathrm{1,c,LPC}$ and $N_\mathrm{2,c,HPC}$ are the map-corrected spool speeds. Further details can be found in \cite{MattiglyBook}. 

Compressor torques $\Gamma_\mathrm{LPC}$ and $\Gamma_\mathrm{HPC}$ are defined as
\begin{align}\label{eq2}
    \Gamma_\mathrm{LPC} &= W_\mathrm{LPC}\,C_{p,\mathrm{a}} \left( \Pi_\mathrm{LPC}^{\frac{\gamma_\mathrm{a}-1}{\gamma_\mathrm{a}}} -1 \right)\frac{1}{\eta_\mathrm{LPC}\,N_1}\,, \\[6pt]
    \Gamma_\mathrm{HPC} &= W_\mathrm{HPC}\,C_{p,\mathrm{a}} \left( \Pi_\mathrm{HPC}^{\frac{\gamma_\mathrm{a}-1}{\gamma_\mathrm{a}}} -1 \right)\frac{1}{\eta_\mathrm{HPC}\,N_2}\,,
\end{align}
where $W$ is the physical air mass flow, $C_{p,\mathrm{a}}$ is the heat capacity of air, and $\gamma_\mathrm{a}$ is the adiabatic exponent of air.

Behind the LPC, the flow splits according to the bypass ratio $BR$ as
\begin{align}
    W_\mathrm{bypass} &= \bigg(\frac{BR}{1 + BR}\bigg) W_\mathrm{LPC}\,, \\[3pt]
    W_\mathrm{HPC} &= W_\mathrm{LPC} - W_\mathrm{bypass}\,.
\end{align}

For the high-pressure turbine (HPT) and low-pressure turbine (LPT) maps, the parameters are read via interpolation as
\begin{equation}
    \begin{aligned}
    \begin{bmatrix}
        \tau_\mathrm{HPT},& W_\mathrm{HPT,c}
    \end{bmatrix}
         &= f(N_\mathrm{2,c,HPT},\Pi_\mathrm{HPT})\,,\\
        \begin{bmatrix}
        \tau_\mathrm{LPT},& W_\mathrm{LPT,c}
    \end{bmatrix} &= f(N_\mathrm{1,c,LPT},\Pi_\mathrm{LPT})\,,
    \end{aligned}
\end{equation}
where $\tau$ denotes the relative total temperature drop $\Delta T_t/T_{t}$, and the corrected quantities are defined in the same way as for compressors, but with turbine inlet parameters \cite{MattiglyBook}.

Turbine torques $\Gamma_\mathrm{HPT}$ and $\Gamma_\mathrm{LPT}$ are defined as
\begin{align}\label{eq_turb_torq}
   \Gamma_\mathrm{HPT} &= W_\mathrm{HPT}\,C_{p,\mathrm{g}}\,T_\mathrm{HPT}\,\tau_\mathrm{HPT}\,\frac{1}{N_2}\,, \\[6pt]
   \Gamma_\mathrm{LPT} &= W_\mathrm{LPT}\,C_{p,\mathrm{g}}\,T_\mathrm{LPT}\,\tau_\mathrm{LPT}\,\frac{1}{N_1}\,,
\end{align}
where $C_{p,\mathrm{g}}$ is the heat capacity of gas and $T_\mathrm{HPT}$ and $T_\mathrm{LPT}$ are turbine inlet total temperatures. 

The combustion chamber (CC) is modeled using the power balance equation 
\begin{equation}\label{eq_CC}
    \eta_\mathrm{CC} W_\mathrm{f} H_\mathrm{L} = W_\mathrm{CC}\, \bar{C}_p(T_\mathrm{CC},T_\mathrm{HPT})(T_\mathrm{HPT} - T_\mathrm{CC}), 
\end{equation}
where $\eta_\mathrm{CC}$, $W_\mathrm{f}$, and $H_\mathrm{L}$ denote the combustion efficiency, fuel mass flow, and lower heating value, respectively, and $\bar{C}_p$, $W_\mathrm{CC}$, $T_\mathrm{CC}$ are the mean heat capacity of gases in CC, CC inlet air mass flow, and CC inlet total temperature, respectively.

The mixer model accounts for the conservation of mass, energy, and momentum. Static pressures at the mixing cross-section must be equal to satisfy the Kutta condition. Given the areas and total parameters in both streams, the static pressure is computed iteratively using adiabatic relations for the mass flow \cite{MattiglyBook}. The following system of equations represents the mixer module:
\begin{equation}
\begin{aligned}
    p_\mathrm{m} &= p_\mathrm{I} = p_\mathrm{II}\,, \\[3pt]
    W_\mathrm{m} &= W_\mathrm{I} + W_\mathrm{II}\,, \\[3pt]
   W_\mathrm{m} C_{p,\mathrm{m}} T_\mathrm{m,t} &= W_\mathrm{I} C_{p,\mathrm{g}} T_\mathrm{I,t} + W_\mathrm{II} C_{p,\mathrm{a}} T_\mathrm{II,t}\,, \\[3pt]
   W_\mathrm{m} v_\mathrm{m} &= p_\mathrm{I} A_\mathrm{m,I} + p_\mathrm{II} A_\mathrm{m,II} - p_\mathrm{m} A_\mathrm{m} \\
   &\quad +\,W_\mathrm{I} v_\mathrm{I} + W_\mathrm{II} v_\mathrm{II}\,, 
\end{aligned}
\end{equation}
with $p$ pressures, $v$ velocities, $A_\mathrm{m} = A_\mathrm{m,I} + A_\mathrm{m,II}$ mixer areas, subscripts $\mathrm{I}$, $\mathrm{II}$, $\mathrm{m}$ denoting mixer outlet, mixer core inlet, and mixer bypass inlet, respectively, and $C_{p,\mathrm{m}}$ the mean mixer heat capacity. 

A convergent nozzle is considered. The nozzle is modeled using a standard adiabatic model with losses, accounting for choked conditions, as presented in detail in \cite{GrasevSpringer2026}. The net thrust is computed as 
\begin{align} \label{eq_thrust}
    F = W_\mathrm{ex} v_\mathrm{ex} - W_\mathrm{LPC} M_0 \sqrt{\gamma_\mathrm{a} R T_0} + A_\mathrm{n}(p_\mathrm{ex} - p_0)\,,
\end{align}
where $W_\mathrm{ex}$ is the exhaust mass flow, $v_\mathrm{ex}$ is the exhaust velocity, $M_0$ is the flight Mach number, $R$ is the universal gas constant, $p_0$ and $T_0$ are the atmospheric pressure and temperature, and $p_\mathrm{ex}$ is the exhaust static pressure, which is equal to $p_0$ when the nozzle is not choked. 

The spool speed dynamics are governed by Newton's second law for rotation, given as
\begin{align}
    \frac{\rm{d}}{\mathrm{d}t}\begin{bmatrix}
        N_1 \\[6pt] N_2
    \end{bmatrix} &= \frac{30}{\pi}\begin{bmatrix}
       \frac{1}{J_1}(\eta_{m,1} \Gamma_\mathrm{LPT} - \Gamma_\mathrm{LPC}) \\[6pt]
       \frac{1}{J_2}(\eta_{m,2} \Gamma_\mathrm{HPT} - \Gamma_\mathrm{HPC})
    \end{bmatrix} \\[6pt] &= \mathbf{F}(N_1,N_2,W_\mathrm{f},A_\mathrm{n},p_\mathrm{1t},T_\mathrm{1t})\,, \label{eq_GTE_F}
\end{align}
where $p_\mathrm{1t}$ and $T_\mathrm{1t}$ are the inlet total pressure and temperature, respectively, capturing the effects of flight conditions, and $J_{1,2}$ and $\eta_{m,1(2)}$ denote the rotor polar moments of inertia and mechanical efficiency of the shafts, respectively.

To perform steady-state and transient computations, a set of nonlinear algebraic equations that enforce mass-flow continuity and thermodynamic consistency is solved using a Newton–Raphson method \cite{GrasevSpringer2026, Yang2024}.

\section{Koopman Operator} \label{sec3}
Koopman operator theory offers an alternative framework for analyzing nonlinear dynamical systems. The main idea is to transform nonlinear dynamics into an infinite-dimensional space of observables that evolve linearly along system trajectories (lifting), and, subsequently, to find a finite-dimensional approximation of the lifted system \cite{Mauroy2020}. Consider the continuous-time autonomous system
\begin{align} \label{eq_sys_1}
    \mathbf{\dot{x} = F(x)},
\end{align}
where $\mathbf{x} \in \mathbb{R}^m$ is the state vector and $\mathbf{F}=\begin{bmatrix} F_1 & F_2 & \cdots & F_m \end{bmatrix}^\intercal$ is a nonlinear drift vector field.

Let $\mathbf{\Psi(x)} \in \mathbb{C}^{n_\Psi}$ denote a nonlinear vector observable with $n_\Psi$ its dimension. The Koopman operator $\mathcal{K}$ acts on observables as \cite{Mezic2012}
\begin{align}
    \mathcal{K}^t\mathbf{\Psi(x) = \Psi \circ F = \Psi}(\mathbf{S}^t(\mathbf{x_0})),
\end{align}
where $\mathbf{S}^t(\mathbf{x_0})$ is the flow map of (\ref{eq_sys_1}) and $\circ$ denotes composition. Hence, nonlinear trajectories are mapped into a linear evolution in the observable space.

The family of operators $\mathcal{K}^t$, parameterized by time $t$, defines the Koopman operator. For discrete-time systems with sampling period $\Delta t$, the operator $\mathcal{K}^{\Delta t}$ governs the evolution from sample to sample as $\mathbf{\Psi(x_{k+1})} = \mathcal{K}^{\Delta t} \mathbf{\Psi(x_k)}$.

This operator family is generated by sampling the infinitesimal Koopman generator $\mathcal{L}$, defined as $\mathcal{L}\mathbf{\Psi} = \lim_{\Delta t \rightarrow 0} (\mathbf{\Psi(x_{k+1})} - \mathcal{K}^{\Delta t} \mathbf{\Psi(x_k)})/\Delta t$ \cite{Mauroy2020, Klus2020}. For simplicity, we retain the notation $\mathcal{K}$.

Since infinite-dimensional Koopman systems are impractical, the Koopman operator is approximated by a finite-dimensional matrix $\mathbf{K}$.

\subsection{Eigenfunctions}
Since $\mathcal{K}$ is linear, it can be represented in terms of its eigenfunctions ${\varphi(\mathbf{x})}\in \mathbf{\Phi}$, with $\mathbf{\Phi}:\mathbb{R}^m \to \mathbb{C}^{n_\varphi}$. The action of $\mathcal{K}$ on $\mathbf{\Phi}$ yields
\begin{align} \label{eq_eig_flow}
    \mathcal{K}^t \mathbf{\Phi(x) = \Phi}(\mathbf{S}^t(\mathbf{x_0})) = e^{\mathbf{\Lambda}t} \mathbf{\Phi({x_0})},
\end{align}
where $\mathbf{\Lambda} \in \mathbb{C}^{n_\varphi \times n_\varphi}$ is a block-diagonal matrix of Koopman eigenvalues.

Considering the approximation matrix $\mathbf{K}$, the eigenfunctions are obtained as
\begin{align} \label{eq_psi_to_phi}
    \mathbf{\Phi = V^\mathrm{-1}\Psi}\,,
\end{align}
where $\mathbf{V}$ is the eigenvector matrix of the Koopman operator obtained from the eigen-decomposition $\mathbf{V\Lambda V^\mathrm{-1} = K}$.

In the time domain, (\ref{eq_eig_flow}) corresponds to the linear system
\begin{align} \label{eq_PhiDot_sys}
    \mathbf{\dot{\Phi}}(t) = \mathbf{\Lambda\Phi}(t), \quad \mathbf{\Phi}(0) = \mathbf{\Phi_0}\,.
\end{align}

Because $\mathbf{\Lambda}$ is diagonal, the eigenfunctions evolve independently under $\mathcal{K}\approx \mathbf{\Lambda}$. Neglecting the case of repeated complex eigenvalues, which is very rare in practice, any general observable $g(\mathbf{x})$ can be projected onto $\mathrm{span}(\mathbf{\Phi})$ using the Koopman mode decomposition \cite{Surana2020}
\begin{align} \label{KMD}
    g(\mathbf{x}(t)) \approx \mathbf{C_g} \mathbf{\Phi}(\mathbf{x}(t)) = \sum_{i=1}^{n_\varphi} c_{g,i}\,\varphi_i(\mathbf{x_0})\,e^{\lambda_i t}\,,
\end{align}
where $\mathbf{C_g}$ is the matrix of Koopman modes. For multiple observables, each row $\mathbf{c_j}$ corresponds to the $j$-th observable $\mathbf{g_j = c_j^\intercal} \mathbf{\Phi}$.

\begin{figure}[!ht]
    \centering
    \includegraphics[width=0.8\linewidth]{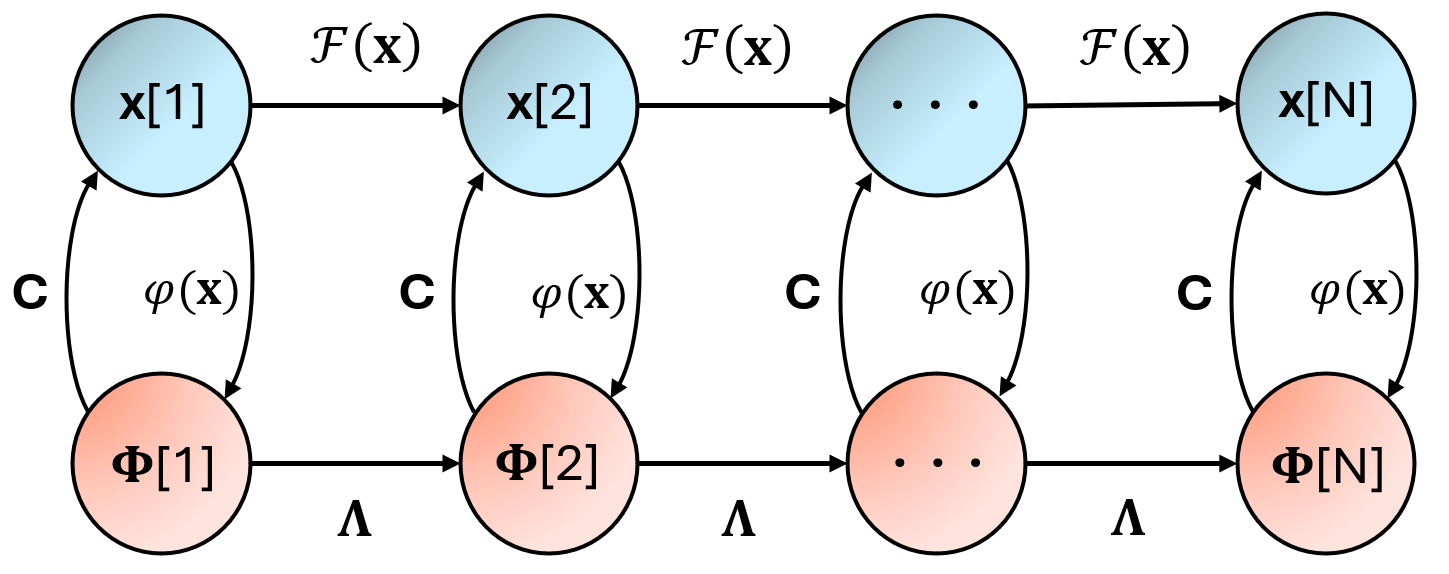}
    \caption{Schematic diagram of Koopman eigenfunction lifting and evolution.}
    \label{Diag_Koopman}
\end{figure}

In practice, the finite subset of eigenfunctions of $\mathcal{K}$ should be a suitable basis for reconstructing observables of interest, e.g., $\mathbf{x \approx C_x} \mathbf{\Phi}$. Fig.~\ref{Diag_Koopman} illustrates the evolution of $\mathbf{\Phi}$ governed by (\ref{eq_PhiDot_sys}) and the reconstruction of the original states via $\mathbf{C_{(x)}}$.

\section{Identification} \label{sec4}
The goal is to obtain a single model suitable for multi-output control design. In this paper, the identification procedure is tailored to ensure an accurate prediction of both spool speeds and EPR for control purposes.

\subsection{Metaheuristic EDMD Algorithm} 
To obtain the Koopman system, the metaheuristic extended dynamic mode decomposition (MH-EDMD) was employed. The approach was originally proposed in \cite{GrasevAccess2025} and builds on EDMD, incorporating L2 regularization and optimizing the nonlinear parameters of observable functions using a metaheuristic method, such as a genetic algorithm or particle swarm optimization. The identification procedure is designed to provide a Koopman model that can be consistently used across multiple control strategies, including mixed-output settings. In this paper, the Koopman system is required to predict not only the dynamics of spool speeds but also the selected output variable: the EPR. Other variables can also be included, e.g., temperatures, pressure ratios, etc.

The Koopman system is given as
\begin{equation}
    \begin{aligned}
        \mathbf{\dot{\Psi}} &= \mathbf{A\Psi + B(x)u}\,,\\[6pt]
        \mathbf{y} &= \mathbf{C \Psi + D u}\,,
    \end{aligned}
\end{equation}
with $\mathbf{x} = [N_1\,\,N_2]^\intercal$ estimated states, $\mathbf{u} = [W_\mathrm{f}\,\,A_\mathrm{n}]^\intercal$ measured inputs, $\mathbf{y} = [\mathbf{x}^\intercal\,\,\Pi_\mathrm{EPR}]^\intercal$ estimated outputs, $\mathbf{A}$ the system matrix, $\mathbf{B(x)}$ the input matrix, and $\mathbf{C = [C_{N1}^\intercal\,\,C_{N2}^\intercal\,\,C_{EPR}^\intercal]^\intercal}$ and $\mathbf{D = [0\,\,D_{EPR}^\intercal]}^\intercal$ are the output matrices.

To reduce the number of decision variables -- parameters, the basis of $\mathbf{B(x)}$ consisted of the observable functions themselves and 1: $\mathbf{B(x) = B_c \,[\mathrm{1}\,\,\Psi(x)^\intercal]^\intercal}$ with $\mathbf{B_c}$ introduced later. 

In EDMD, the states $\mathbf{x}$ are first lifted using the observables $\mathbf{\Psi}$ and concatenated with the inputs as
\begin{align}
    \mathbf{Y = [\Psi^\intercal \,\, u^\intercal \,\,\Psi^\intercal}u_1 \,\, \mathbf{\Psi}^\intercal u_2]^\intercal \in \mathbb{R}^{(3n_\Psi+2) \times N_t}\,,
\end{align}
where $n_\Psi$ is the number of observables and $N_t=t_\mathrm{end}/\Delta t$ is the number of samples with $\Delta t$ the sampling period.

Subsequently, the observable dataset is split into two matrices: 
\begin{equation}
    \begin{aligned}
        \boldsymbol{\mathcal{Y}_1} &= [\mathbf{Y_1\,\, Y_2\,\,\cdots\,\, Y_{N_t-1}} ]\,,\\[3pt]
        \boldsymbol{\mathcal{Y}_2} &= [\mathbf{Y_2\,\, Y_3\,\,\cdots\,\, Y_{N_t}} ] = \mathcal{S}\boldsymbol{\mathcal{Y}_1}\,,
    \end{aligned}
\end{equation}
with $\mathcal{S}$ denoting the shift operator. 

The Koopman operator is approximated using regularized least squares as
\begin{align} \label{eq_K}
    \mathbf{K^\intercal = (\boldsymbol{\mathcal{Y}_1\mathcal{Y}_1}^\intercal + \alpha I)^\mathrm{-1} \boldsymbol{\mathcal{Y}_1 \mathcal{Y}_2}^\intercal}\,,    
\end{align}
where $\alpha$ is the regularization parameter. 

The matrices $\mathbf{A}$ and $\mathbf{B_c}$ are obtained by first extracting the discrete-time matrices $\mathbf{A_d}$ and $\mathbf{B_d}$ from $\mathbf{K}$ as
\begin{align}
    \begin{bmatrix}
        \mathbf{A_d}_{\,n_\Psi \times n_\Psi} & \mathbf{B_d}_{\,n_\Psi \times (2n_\Psi+2)} \\
        \cdot & \cdot
    \end{bmatrix} = \mathbf{K}
\end{align}
and converting to the continuous-time form using $\mathbf{A = \log (A_d)} / \Delta t \approx \mathbf{(A_d - I)}/\Delta t$ and $\mathbf{B_c \approx B_d} / \Delta t$, which is a reasonable approximation for small $\Delta t$.

The time-varying input dynamics are constructed as
\begin{align}
    \mathbf{B(x) = [B_{u1} + B_{xu1}\Psi(x) \,\,\,\, B_{u2} + B_{xu2}\Psi(x)}]\,,
\end{align}
where $\mathbf{[B_{u1}\,\,B_{u2}\,\,B_{xu1}\,\,B_{xu2}] = B_c}$.

Given that the first two observables are the spool speeds, $\psi_1=N_1$ and $\psi_2=N_2$, the corresponding spool-speed output matrix is given as $\mathbf{C_N = [I\,\,0\,\,\cdots\,\,0]}$ and the EPR output matrices are obtained as $[\mathbf{C_{EPR} \,\, D_{EPR}}] = \Pi_\mathrm{EPR}\,\mathbf{[\Psi\,\,u]}^\dagger$. This solution and (\ref{eq_K}) are the minimum-norm solutions as discussed and proved in \cite{Chen2024-MinNormProof}.

The objective function minimizes prediction error across the entire training and validation time series, computed by numerically integrating the Koopman system, rather than relying solely on the one-step prediction error, as in classical EDMD. In this way, the accumulation of prediction error is explicitly captured, increasing the robustness of the MH-EDMD. In this paper, the mean absolute error (MAE) was selected as the main cost function for prediction. This accumulated error is difficult to address using gradient-based optimization; hence, the metaheuristic approach.

Furthermore, the metaheuristic nature allows for explicitly penalizing lightly unstable eigenvalues that can occur even in solutions with good prediction accuracy and might hamper long-term prediction. For this, the stability margin is specified as the maximum real part of the dominant eigenvalue of the system matrix. A lower bound on the real part is also imposed to ensure that all modes contribute to the dynamics, discarding those that decay too rapidly.

The objective function also penalizes uncontrollable and unobservable solutions, utilizing the extended Kalman rank condition for controllability (for details, see \cite{GrasevAccess2025}) and classical rank condition for observability \cite{Friedland1987}. The final optimization problem is thus
\begin{equation} \label{eq_KOGA}
    \begin{aligned}
        &\min_{\mathbf{p}} {\frac{1}{N_t}\sum_{k = 1}^{N_t} \big|\mathbf{y_k - C\Psi_k(p)}\big|} \\[6pt]
        &\mathrm{s.t.} \quad \mathrm{Re}(\mathbf{\Lambda})_\mathrm{min} \le \mathrm{Re}(\mathbf{\Lambda}) \le \mathrm{Re}(\mathbf{\Lambda})_\mathrm{max}\,,  \\[6pt]
        &\qquad\,\, \mathrm{rank}\,[\mathbf{B_c\,\,AB_c \,\, \cdots \,\, A^\mathrm{n_\Psi-1}B_c}] = n_\Psi\,, \\[6pt]
        &\qquad\,\, \mathrm{rank}\,{[\mathbf{C^\intercal\,\,(C A)^\intercal \,\, \cdots \,\, (C A^\mathrm{n_\Psi-1})^\intercal}]}^\intercal = n_\Psi\,, \\[6pt]
        &\qquad\,\, \inf \boldsymbol{\mathcal{D}}(\mathbf{p}) \ge d_\mathrm{min}\,,
    \end{aligned}
\end{equation}
where $\mathbf{p} \in \mathcal{P}$ is a vector of parameters of the lifting functions and $\boldsymbol{\mathcal{D}}$ is an upper-triangular matrix, measuring the Euclidean distance of observables in the parameter space $\mathcal{P}$, with elements $d_{ij} = \|\mathbf{p_i - p_j}\|_2$, $i=1,...,n_\Psi-3$, $j=i+1,...,n_\Psi-2$, and the remaining elements replaced by large numbers. This last condition promotes diversity.

As a result, the optimization balances the prediction accuracy with the system's dynamical properties, ensuring stability, controllability, and observability of the Koopman model. 

After the matrix $\mathbf{A}$ is obtained, its eigen-decomposition is used to transform the identified system into an invariant eigenfunction system in the Jordan canonical form using (\ref{eq_psi_to_phi}) as
\begin{equation}\label{eig_sys_conv}
    \begin{aligned}
          \mathbf{\dot{\Phi}} &= \mathbf{\Lambda\Phi + G(x)u} \,, \\
          \mathbf{y} &= \mathbf{C_\Phi \Phi + D u} \,, \\
    \end{aligned}
\end{equation}
where $\mathbf{\Lambda = V^\mathrm{-1} A V}$, $\mathbf{G(x) = V^\mathrm{-1}B(x)}$, and $\mathbf{C_\Phi = C V}$.

The algorithm \ref{alg_MH_EDMD} summarizes the MH-EDMD.
\begin{algorithm}[!ht] 
\caption{MH-EDMD Identification}
\KwIn{Data $\{\mathbf{x,u,y}\}$, $\Delta t$, $\alpha$}
\KwOut{Koopman model $(\mathbf{\Lambda, G(x), C_\Phi,D})$}
Initialize parameter population $\mathcal{P}$\;
\While{not converged}{
    \ForEach{$\mathbf{p} \in \mathcal{P}$}{
        Define observables $\mathbf{\Psi(x;p})$\;

        Construct lifted data $\mathbf{Y}$\;

        Form $\boldsymbol{\mathcal{Y}_1}$, $\boldsymbol{\mathcal{Y}_2}$\;

        Compute $\mathbf{K}$ via regularized least squares\;

        Extract $\mathbf{A_d}, \mathbf{B_d}$ and convert to $\mathbf{A}, \mathbf{B_c}$\;

        Build $\mathbf{B(x)}$, $\mathbf{C}$, $\mathbf{D}$\;

        Simulate system and compute MAE\;

        Evaluate constraints (stability, controllability, observability, diversity)\;
    }
    Update $\mathcal{P}$ using GA/PSO\;
}
Transform to eigenfunction form\;
\label{alg_MH_EDMD}
\end{algorithm}

\subsection{Actuator Dynamics}
To account for actuator dynamics, the Koopman system (\ref{eig_sys_conv}) was augmented with the actuator system
\begin{align}\label{act_eqn}
    \underbrace{\begin{bmatrix} \dot{W}_\mathrm{f} \\[6pt] \dot{A}_\mathrm{n}\end{bmatrix}}_{\mathbf{\dot{x}_{act}}} = 
    \underbrace{\begin{bmatrix} -\frac{1}{T_\mathrm{f}} & 0 \\ 0 & -\frac{1}{T_\mathrm{a}}
    \end{bmatrix}}_{\mathbf{A_{act}}} \underbrace{\begin{bmatrix} W_\mathrm{f} \\[6pt] A_\mathrm{n}
    \end{bmatrix}}_{\mathbf{x_{act}}} + \underbrace{\begin{bmatrix}
        \frac{1}{T_\mathrm{f}} &\ 0 \\ 0 &\ \frac{1}{T_\mathrm{a}} \end{bmatrix}}_{\mathbf{B_{act}}}
    \underbrace{\begin{bmatrix}
        W_\mathrm{f}^{c} \\[6pt] A_\mathrm{n}^{c}
    \end{bmatrix}}_{\mathbf{u}},
\end{align}
where $T_\mathrm{f} = 0.06$ s and $T_\mathrm{a} = 0.1$ s are the fuel system and nozzle time constants, respectively, and superscript $(\cdot)^{c}$ denotes the controller commands, new inputs to the augmented system.

The resulting augmented system is
\begin{equation}
\begin{aligned} \label{eq_aug}
\underbrace{\begin{bmatrix}
    \mathbf{\dot{\Phi}} \\ \mathbf{\dot{x}_{act}}
\end{bmatrix}}_{\mathbf{\dot{z}}} &=
\underbrace{\begin{bmatrix}
        \mathbf{\Lambda} & \mathbf{G}\mathbf{(x)}\\
        \mathbf{0} & \mathbf{A_{act}} 
    \end{bmatrix}}_{\mathbf{A_{z}(x)}}
\underbrace{\begin{bmatrix}
    \mathbf{{\Phi}} \\ \mathbf{x_{act}}
\end{bmatrix}}_{\mathbf{z}} + 
\underbrace{\begin{bmatrix}
    \mathbf{0} \\ \mathbf{B_{act}}
\end{bmatrix}}_{\mathbf{B_{z}}}
\mathbf{u}\,, \\[6pt]
\mathbf{y} &= \underbrace{\begin{bmatrix}
    \mathbf{C_\Phi} & \mathbf{D} \end{bmatrix}}_{\mathbf{C_{z}}}\,\mathbf{z}\,, 
\end{aligned}
\end{equation}
where $\mathbf{z}$ is the augmented state and $\mathbf{C_z}$ will now on be denoted just $\mathbf{C}$ for simplicity.

\subsection{Kalman Filter Design} \label{sec_KF}
The eigenfunction states cannot be measured. Since the identification process ensures that the system is observable, a Kalman filter can be designed in the lifted state space to estimate eigenfunctions and improve prediction accuracy. Because the outputs are accurately reconstructed using the lifted states, improving the estimation of lifted states also improves the output estimation. 

In the Kalman filter design, the actuator dynamics are excluded because the actuator states are known and do not need to be estimated. Consider the eigenfunction system (\ref{eig_sys_conv}) with noise:
\begin{equation}
\begin{aligned}\label{eq_noisy_sys}
    &\mathbf{\dot{\Phi} = \Lambda\Phi + \mathbf{G} x_{act} + w_p}, \quad \mathbf{w_p} \sim \mathcal{N}(\mathbf{0, \Sigma_p}), \\[3pt]
    &\mathbf{y = C_\Phi \Phi + Dx_{act} + w_n}, \quad \mathbf{w_n} \sim \mathcal{N}(\mathbf{0, \Sigma_n}), 
\end{aligned}
\end{equation}
where $\mathbf{w_p}$ and $\mathbf{w_n}$ are the process and measurement noise, respectively, both assumed to be Gaussian white noise, and $\mathbf{\Sigma_n}$ and $\mathbf{\Sigma_p}$ are the noise covariance matrices. 

Since $\mathbf{\Lambda}$ and $\mathbf{C_\Phi}$ are constant, the optimal Kalman filter gain matrix $\mathbf{L}$ can be obtained using the observer algebraic Riccati equation as follows:
\begin{equation} \label{eq_KF_L}
    \mathbf{L = P_o C_\Phi^\intercal R_o^\mathrm{-1}},
\end{equation}
where $\mathbf{P_o \succ 0}$ is a solution of the observer Riccati equation \cite{Friedland1987}:
\begin{equation} \label{eq_KF_Riccati}
    \mathbf{\Lambda P_o + P_o \Lambda^\intercal + G_o Q_o G_o^\intercal - P_o C_\Phi^\intercal R_o^\mathrm{-1} C_\Phi P_o = 0},
\end{equation}
where $\mathbf{G_o}$ is the process noise gain matrix usually set to identity $\mathbf{I}$, and $\mathbf{R_o \succ 0}$ and $\mathbf{Q_o \succeq 0}$ are the measurement and process noise covariance matrices, respectively, estimated from the data.

\section{Considered Controllers} \label{sec5}

\subsection{Koopman Adaptive Model Predictive Control}
An adaptive Koopman MPC is considered in this paper. In the adaptive KMPC, the time-varying Koopman model is evaluated at the current time step $k$ and held constant over the prediction horizon. This approximation preserves the linear structure of the prediction model and enables efficient quadratic programming.

Consider a discrete-time version of the augmented system (\ref{eq_aug}) given as
\begin{align}
    \mathbf{z_{k+1} = \boldsymbol{\mathcal{L}}_k\,z_k + \Gamma u_k}\,,
\end{align}
where $\mathbf{\boldsymbol{\mathcal{L}}_k = A_{z}}(k\Delta t)\Delta t + \mathbf{I}$ and $\mathbf{\Gamma = B_{z}}\Delta t$.

To improve tracking performance, the system is reformulated in an incremental form. To obtain an incremental formulation, consider
\begin{equation}
\begin{aligned}
    \mathbf{z_{k+1}} &= \mathbf{\boldsymbol{\mathcal{L}}_k z_k + \Gamma u_k}\,, \\[3pt]
    \mathbf{z_{k}} &= \mathbf{\boldsymbol{\mathcal{L}}_{k-1} z_{k-1} + \Gamma u_{k-1}}\,.
\end{aligned}
\end{equation}


Here, the time-varying matrix $\boldsymbol{\mathcal{L}}_k$ is kept constant over the prediction horizon $n_p$. Therefore, with a slight abuse of notation, subtraction of the equations yields
\begin{equation}
\begin{aligned}
    \underbrace{\mathbf{z_{k+1} - z_k}}_{\mathbf{\Delta z_{k+1}}}
    &= \mathbf{\boldsymbol{\mathcal{L}}_k\underbrace{\mathbf{(z_k - z_{k-1})}}_{\mathbf{\Delta z_k}} + \Gamma\underbrace{\mathbf{(u_k - u_{k-1})}}_{\mathbf{\Delta u_k}}}\,, \\
    \underbrace{\mathbf{y_{k+1} - y_k}}_{\mathbf{\Delta y_{k+1}}}
    &= \mathbf{C\,\Delta z_{k+1}}\,.
\end{aligned}
\end{equation}

To achieve reference tracking, the incremental state is augmented with the output as $\mathbf{x_a = [\Delta z^\intercal \ \ y^\intercal]^\intercal}$. The augmented system is then given by
\begin{align} \label{eq_aug_evolution}
    \underbrace{\begin{bmatrix}
        \mathbf{\Delta z_{k+1}} \\[3pt] \mathbf{y_{k+1}}
    \end{bmatrix}}_{\mathbf{x_{a,k+1}}} = 
    \underbrace{\begin{bmatrix}
        \mathbf{\boldsymbol{\mathcal{L}}_k} & \mathbf{0} \\[3pt]
        \mathbf{C\boldsymbol{\mathcal{L}}_k} & \mathbf{I}
    \end{bmatrix}}_{\mathbf{A_{a,k}}} 
    \underbrace{\begin{bmatrix}
        \mathbf{\Delta z_{k}} \\[3pt] \mathbf{y_{k}}
    \end{bmatrix}}_{\mathbf{x_{a,k}}} + 
    \underbrace{\begin{bmatrix}
        \mathbf{\Gamma} \\[3pt] \mathbf{C\Gamma}
    \end{bmatrix}}_{\mathbf{B_a}} \mathbf{\Delta u_k}\,.
\end{align}

The AKMPC algorithm minimizes a quadratic cost function at time step $k$ as
\begin{align} \label{cost1}
    \mathcal{J}_k
    &= \|\mathbf{e_{y,k+n_p}}\|_\mathbf{Q_T}^2
    + \sum_{j=k}^{k+n_p-1}\left(\|\mathbf{e_{y,j}}\|^2_{\mathbf{Q_y}} + \|\mathbf{\Delta u_j}\|_\mathbf{R}^2\right)\,,
\end{align}
where $\|(\cdot)\|_\mathbf{P}^2 = {(\cdot)}^\intercal\mathbf{P}(\cdot)$, $\mathbf{e_y = y - y_{ref}}$ is the tracking error, $\mathbf{Q_y\succ0}$, $\mathbf{R\succ0}$, and $\mathbf{Q_T\succeq 0}$ are the stage tracking error, input-increment, and terminal cost weighting matrices, respectively.

To reduce the number of decision variables, the input increment is held constant after the control horizon $n_c \le n_p$, i.e., $\mathbf{\Delta u_j = \Delta u_{k+n_c-1}}$ for $j = k+n_c,\dots,k+n_p-1$.

The stacked output prediction can then be written as
\begin{equation}  
\begin{aligned} \label{eq_O_M_form_Nc}
    &\underbrace{\begin{bmatrix}
        \mathbf{y_{k+1}} \\ \mathbf{y_{k+2}} \\ \vdots \\ \mathbf{y_{k+n_p}}
    \end{bmatrix}}_{\mathbf{Y_k}}
    =
    \underbrace{\begin{bmatrix}
        \mathbf{C_a A_{a,k}} \\ \mathbf{C_a A_{a,k}^2} \\ \vdots \\ \mathbf{C_a A_{a,k}^{n_p}}
    \end{bmatrix}}_{\mathbf{O_k}}
    \mathbf{x_{a,k}} \\
    &+
    \underbrace{\begin{bmatrix}
        \mathbf{C_a B_a} & \mathbf{0} & \cdots & \mathbf{0} \\
        \mathbf{C_a A_{a,k} B_a} & \mathbf{C_a B_a} & \cdots & \mathbf{0} \\
        \vdots & \vdots & \ddots & \vdots \\
        \mathbf{C_a A_{a,k}^{n_c-1} B_a} & \mathbf{C_a A_{a,k}^{n_c-2} B_a} & \cdots & \mathbf{C_a B_a} \\
        \mathbf{C_a A_{a,k}^{n_c} B_a} & \mathbf{C_a A_{a,k}^{n_c-1} B_a} & \cdots & \mathbf{S_{n_c+1}} \\
        \vdots & \vdots & \ddots & \vdots \\
        \mathbf{C_a A_{a,k}^{n_p-1} B_a} & \mathbf{C_a A_{a,k}^{n_p-2} B_a} & \cdots & \mathbf{S_{n_p}}
    \end{bmatrix}}_{\mathbf{M_k}}
    \underbrace{\begin{bmatrix}
        \mathbf{\Delta u_k} \\ \mathbf{\Delta u_{k+1}} \\ \vdots \\ \mathbf{\Delta u_{k+n_c-1}}
    \end{bmatrix}}_{\mathbf{\Delta U_k}},
\end{aligned}
\end{equation}
where $\mathbf{C_a = [0 \ \ I]}$ and
\begin{align}
\mathbf{S_j}
= \sum_{i=0}^{j-n_c-1}\mathbf{C_a A_{a,k}^{i}B_a},
\qquad j = n_c+1,\dots,n_p\,.
\end{align}

The predicted tracking error is then
\begin{align} \label{eq_MPC_err}
    \mathbf{E_k} = \mathbf{Y_k} - \mathbf{Y_{ref,k}},
\end{align}
where $\mathbf{Y_{ref,k}}$ is a stacked vector of the reference outputs. In this paper, $\mathbf{y_{ref,j} = y_{ref,k}}$ for $j=k+1,\dots,k+n_p$, since future pilot commands cannot be predicted and this is a reasonable assumption for small $n_p$.

The stacked output and input weighting matrices are given as
\begin{align}
    \mathbf{\bar{Q}} &= \mathbf{I}_{n_p\times n_p} \otimes \mathbf{Q_{y}}\,, \\[3pt]
    \mathbf{\bar{R}} &= \mathbf{I}_{n_c\times n_c} \otimes \mathbf{R}\,, 
\end{align}
where $\otimes$ denotes the Kronecker product.

\subsubsection{Constraints} \label{sec_cons}

\paragraph{Inputs}
The constraints on the absolute values of inputs in the sequence are imposed by expressing the inputs as a sum of the successive increments, $\mathbf{u_{k+j} = u_{k-1} + \sum_{i=0}^j {\Delta u_{k+i}}}$. Thus, the matrices of the corresponding constraints are 
\begin{align}
    \mathbf{\mathcal{A}_u} &= \begin{bmatrix}
        \mathbf{T} \\ -\mathbf{T}
    \end{bmatrix}\,, \quad \mathbf{b_{u,k}} = \begin{bmatrix}
        \mathbf{U_{max,k}- U^p_k}  \\[3pt] -(\mathbf{U_{min,k} - U^p_k)}
    \end{bmatrix}\,,
\end{align}
where
\begin{align}
    \mathbf{T} &= \begin{bmatrix}
        \mathbf{I}_{m\times m} & \mathbf{0} & \cdots & \mathbf{0} \\
        \mathbf{I}_{m\times m} & \mathbf{I}_{m\times m} & \cdots & \mathbf{0} \\
        \vdots & \vdots & \ddots & \vdots \\
        \mathbf{I}_{m\times m} & \mathbf{I}_{m\times m} & \mathbf{I}_{m\times m} & \mathbf{I}_{m\times m} \end{bmatrix}\,, 
\end{align}
for the absolute values, where $\mathbf{U^p}$, $\mathbf{U_{max}}$, and $\mathbf{U_{min}}$ denote the stacked vectors of the previous-step applied input and input limits, respectively, and 
\begin{align}
    \mathbf{\mathcal{A}_{\Delta u}} &= \begin{bmatrix}\mathbf{I}_{mn_c \times mn_c} \\[3pt] -\mathbf{I}_{mn_c \times mn_c}\end{bmatrix}\,, \quad 
    \mathbf{b_{\Delta u}} = \begin{bmatrix} \mathbf{\Delta U_{max}} \\[3pt] \mathbf{-\Delta U_{min}}\end{bmatrix} 
\end{align}
for the input increments.

\paragraph{Outputs and GTE Safety Limiters}
For safe GTE operation, it is crucial to avoid exceeding the limits of some outputs. Particularly important is avoiding compressor surge, combustor blowout, and exceeding high-pressure turbine inlet temperature (HPT TIT) \cite{Jaw2009}.  

The lean blowout limit is a function of the fuel-air ratio and can be addressed via input constraints. Similarly, the other limits can also be addressed via input constraints \cite{GrasevSpringer2026}. To further enhance compressor surge avoidance, the acceleration rate $\dot{N}$ can be limited, since the operating point on the compressor map moves towards the surge line during sharp transients \cite{Jaw2009}. 

Usually, PID controllers and min-max logic are used inside the GTE limiters, acting on the inputs at the current control instance. This can lead to sharp transitions between the limiters, causing abrupt changes in the fuel flow and potential violation of the limits. In the MPC, this can be avoided thanks to its predictive nature.

The $\dot{N}$ and TIT are highly nonlinear functions of states and inputs. In classical MPC approaches, the linearized output maps are employed, which may fail to capture the nonlinear relations during large transients, possibly leading to overly conservative control. This motivates the use of the Koopman framework, in which nonlinear output relations can be projected onto the span of the observables and inputs, yielding potentially more accurate linear predictions. Therefore, these limiters can now be expressed as linear constraints in the AKMPC.
 
The LPC is more susceptible to surge. The over-temperature directly corresponds to the TIT, and lean blowout limits can be represented by TIT and acceleration rate limits. Thus, the low-pressure rotor acceleration rate, $\dot{N}_1$, and the TIT are selected as examples and can be expressed as
\begin{align}
    \dot{N}_1 &= \mathbf{C_{N1} \Lambda \Phi} + \mathbf{C_{N1} G x_{act}}  = \mathbf{C_{\dot{N}1} z}\,, \\[6pt]
    \mathrm{TIT} &= \mathbf{C_{TIT,\Phi} \Phi + C_{TIT,act} x_{act}} = \mathbf{C_{TIT} z}\,.
\end{align}

The prediction of $\dot{N}_1$ is given as
\begin{align}
    \begin{bmatrix}
        \dot{N}_{1,k} \\ \vdots \\ \dot{N}_{1,k+n_p}
    \end{bmatrix} &= \mathbf{O_{\dot{N}1,k} z_k} + \mathbf{M_{{\dot{N}1},k}(U^p + T \Delta U)_k}\,,
\end{align}
with the matrices $\mathbf{O_{N1}}$ and $\mathbf{M_{N1}}$ constructed in the same fashion as for state prediction.

The TIT prediction follows the same structure, and the $\dot{N}_1$ and TIT constraints can be augmented in a final output constraint as
\begin{equation}
    \mathbf{\mathcal{A}_{y,k} \Delta U_k \le b_{y,k}}
\end{equation}
with
\begin{equation}
    \mathbf{\mathcal{A}_{y,k}} = \begin{bmatrix}
        \mathbf{M_{{\dot{N}1},k} T} \\ -\mathbf{M_{{\dot{N}1},k} T} \\ \mathbf{M_{TIT,k} T}
    \end{bmatrix}
\end{equation}
and
\begin{equation} \label{eq_ineq_RHS}
    \mathbf{b_{y,k}} = \begin{bmatrix}
        \mathbf{\dot{N}_{1,max} - O_{\dot{N}1,k} z_k - M_{{\dot{N}1},k} U^p_k} \\ -\mathbf{\dot{N}_{1,min} + O_{\dot{N}1,k} z_k + M_{{\dot{N}1},k}U^p_k} \\ \mathbf{TIT_{max} - O_{TIT,k} z_k - M_{TIT,k}U^p_k}
    \end{bmatrix}\,.
\end{equation}

It should be noted that the prediction for the limited parameters will inevitably be burdened by error and uncertainty. Therefore, the constraints for the predicted quantities should be tighter than the physical limits. Also, the minimum TIT can be included to address the lean blowout limit together with $\dot{N}_\mathrm{1, min}$.

Other outputs, e.g., $\dot{N}_2$, fuel-air-ratio, or spool speeds, can be included as well. The augmentation with the input constraints yields the final constraints $\mathbf{\mathcal{A}_k \Delta U_k \le b_k}$.

\subsubsection{Disturbance Observer}
Suppose that the measured output $\mathbf{y}$ differs from the estimated output $\mathbf{\hat{y}}$ due to model mismatch represented by disturbances, i.e., $\mathbf{y = \hat{y} + d}$. 


To account for this model mismatch during control and improve the prediction accuracy of outputs, especially the normalized spool speeds used for evaluation of the time-varying Koopman system, while lowering the effects of measurement noise, the system can be augmented with a disturbance observer:
\begin{align}
    \mathbf{\hat{d}_{k+1} = \hat{d}_k + L_d\,i_k}\,,
\end{align}
where $\mathbf{L_d}$ is the DO gain matrix and $\mathbf{i_k = y_k - C z_k}$ is the innovation term.

The output prediction is subsequently corrected using the estimated disturbance as $\mathbf{\tilde{y} = \hat{y} + \hat{d}}$. 

Since the disturbances vary slowly over time, a steady-state Kalman filter can be employed. The gain matrix $\mathbf{L}$ is obtained by solving discrete-time versions of equations (\ref{eq_KF_L}) and (\ref{eq_KF_Riccati}). The observer design procedure follows that of the Kalman filter in Section \ref{sec_KF}. The actuator states are known and measurable, and do not have to be estimated. Thus, only the system (\ref{eig_sys_conv}) is augmented with the disturbance states, yielding the following system:
\begin{equation}
\begin{aligned} 
    \begin{bmatrix}
        \mathbf{\Phi_{k+1}} \\[3pt] \mathbf{d_{k+1}}
    \end{bmatrix} &= 
    \underbrace{\begin{bmatrix}
        \mathbf{\Lambda_d} & \mathbf{0} \\[3pt]
        \mathbf{0} & \mathbf{I}
    \end{bmatrix}}_{\mathbf{A_{e}}} \begin{bmatrix}
        \mathbf{\Phi_{k}} \\ \mathbf{d_{k}}
    \end{bmatrix} + \underbrace{\begin{bmatrix}
        \mathbf{G_{d,k}} \\[3pt] \mathbf{0}
    \end{bmatrix}}_{\mathbf{B_{e,k}}} \mathbf{x_{act,k}} + \underbrace{\begin{bmatrix}
        \mathbf{L_\Phi} \\[3pt] \mathbf{L_d}
    \end{bmatrix}}_{\mathbf{L}} \mathbf{i_k}\,, \\[0pt]
    \mathbf{\tilde{y}_{k+1}} &= \underbrace{\begin{bmatrix}
        \mathbf{C_\Phi} & \mathbf{I}
    \end{bmatrix}}_{\mathbf{C_e}} \begin{bmatrix}
        \mathbf{\Phi_{k+1}} \\[3pt] \mathbf{d_{k+1}}
    \end{bmatrix} + \mathbf{D\,x_{act,k+1}}\,,
\end{aligned}
\label{eq_aug_disturbance}
\end{equation}
where $\mathbf{\Lambda_d} = \mathrm{exp}(\mathbf{\Lambda} \Delta t)$, $\mathbf{G_d = G}\Delta t$, and $\mathbf{A_e}$, $\mathbf{B_e}$, and $\mathbf{C_e}$ are the augmented system matrix, input matrix, and output matrix of the estimator, respectively. 

Substituting (\ref{eq_MPC_err}), with $\mathbf{Y_k}$ computed using (\ref{eq_O_M_form_Nc}), into the cost function yields the following quadratic program (QP) at each time step $k$:
\begin{equation}
\begin{aligned} \label{eq_final_MPC_problem}
    &\min_{\mathbf{\Delta U_k}} \left(\mathbf{\Delta U_k^\intercal H_k \Delta U_k + f_k^\intercal \Delta U_k}\right) \quad s.t. \\[6pt]
    &\mathbf{\mathcal{A}_k \Delta U_k \le b_k}\,, 
\end{aligned}
\end{equation}
where $\mathbf{H_k \succ 0}$ and $\mathbf{f_k}$ are the Hessian and the linear term of the cost function, respectively, given as
\begin{align}
    \mathbf{H_k} &= \mathbf{M_k^\intercal\bar{Q}M_k + \bar{R}}\,,\\[6pt]
    \mathbf{f_k} &= \mathbf{M_k^\intercal\,\bar{Q}\,(O_k\, x_{a,k} - Y_{ref,k})}\,.
\end{align}

Once an optimal solution $\mathbf{\Delta U_k^*}$ is obtained, only the first input increment $\mathbf{\Delta u_k^*}$ is used for the control as
\begin{align}
    \mathbf{u_{k} = u_{k-1} + \Delta u_k^*}\,.
\end{align}

\begin{figure}[!ht]
    \centering
    \includegraphics[width=1\linewidth]{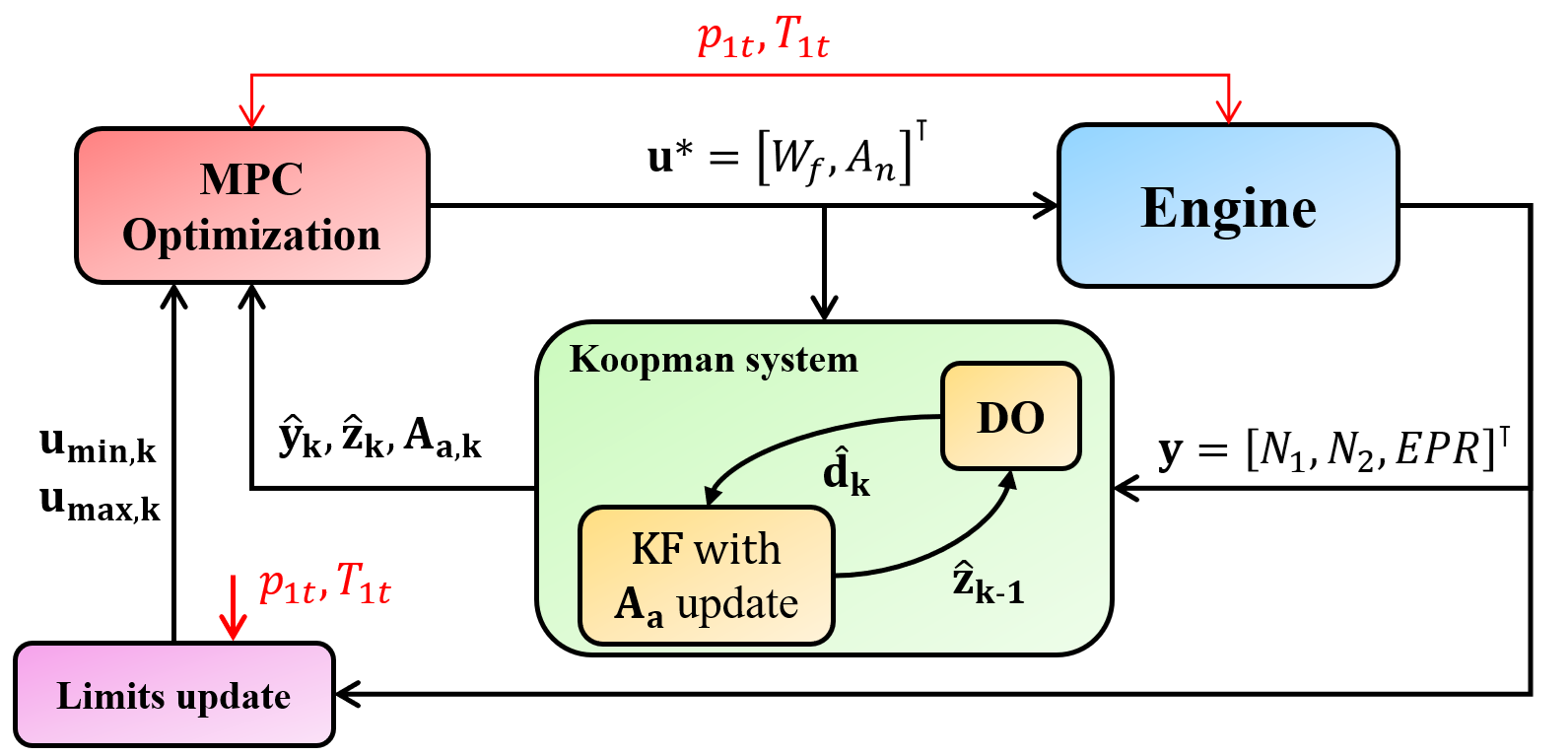}
    \caption{The AKMPC diagram showing the interaction between the Koopman system and MPC optimization.}
    \label{AKMPC_Diagram}
\end{figure}

The constrained optimization problem (\ref{eq_final_MPC_problem}) remains convex, allowing for efficient solution via QP algorithms. The AKMPC controller diagram is depicted in Fig. \ref{AKMPC_Diagram}.

\subsection{Feedback Linearization Using a Koopman-Derived Model}

To showcase a broader applicability of the Koopman model to other nonlinear control methods, a feedback linearization controller is also proposed, serving as a nonlinear control benchmark for comparison to the AKMPC.

Using the affine linear-parameter varying Koopman system, a control-affine nonlinear model can be derived in the original state space for the design of a feedback linearization controller. Considering the Koopman model (\ref{eig_sys_conv}), the derivation for spool speeds reads
\begin{equation}
\begin{aligned} \label{eq_FBL_Spool_Speeds}
    \mathbf{\dot{x}} &= \mathbf{C_{\Phi N} \dot{\Phi}} \\ 
    &= \mathbf{C_{\Phi N}\big(\Lambda\Phi(x) + G(x) x_{act}\big)}\,, \\[3pt]
    \mathbf{\ddot{x}} &= \mathbf{C_{\Phi N}\big(\Lambda\nabla\Phi} + x_\mathrm{act,1}\mathbf{\nabla G_1} + x_\mathrm{act,2}\mathbf{\nabla G_2\big)\,\dot{x}}\\
    & \quad + \mathbf{C_{\Phi N} G(x) \dot{x}_{act}}\,,
\end{aligned}
\end{equation}
where $\mathbf{C_{\Phi N}}$ are the first two rows of $\mathbf{C_\Phi}$ and $\mathbf{\nabla \Phi}$, $\mathbf{\nabla G_1}$, and $\mathbf{\nabla G_2}$ denote Jacobians of the eigenfunction vector $\mathbf{\Phi(x)}$ and of the first and second columns of the input mapping $\mathbf{G(x)}$. 

For the EPR, the derivation reads
\begin{equation} \label{eq_FBL_EPR}
    \begin{aligned}
        \dot{\Pi}_\mathrm{EPR} = \mathbf{C_{EPR} \big(\Lambda\Phi(x) + G(x) x_{act}\big) + D_{EPR}\dot{x}_{act}}\,.
    \end{aligned}
\end{equation}

Substituting the actuator dynamics (\ref{act_eqn}), equations (\ref{eq_FBL_Spool_Speeds}) and (\ref{eq_FBL_EPR}) can be converted to a control-affine system
\begin{equation}    
\begin{aligned}
    \mathbf{\ddot{x}} &= \mathbf{\mathcal{F}_{x}(x,x_{act}) + \mathcal{G}_{x}(x,x_{act})u}\,, \\[6pt]
    \dot{\Pi}_\mathrm{EPR} &= \mathbf{\mathcal{F}_{EPR}(x,x_{act}) + \mathcal{G}_{EPR}(x,x_{act}) u}\,,
\end{aligned}
\end{equation}
with $\mathcal{F}$ and $\mathcal{G}$ denoting the drift and input dynamics, respectively. These functions follow directly from (\ref{eq_FBL_Spool_Speeds}) and (\ref{eq_FBL_EPR}).

In the final implementation, for a given choice of two controlled outputs, the corresponding output time derivative equations are stacked to form the drift vector field $\mathcal{F}\in\mathbb{R}^{2\times 1}$ and a square decoupling matrix $\mathcal{G}\in\mathbb{R}^{2\times 2}$. Further assume $\mathcal{G}$ is invertible, and consider an input vector generally given as 
\begin{align}
    \mathbf{u = \mathcal{G}^{\mathrm{-1}} \big(v - \mathcal{F}\big)}\,,
\end{align}
where $\mathbf{v}$ is a virtual input.

For the system above, the closed-loop dynamics are given as $\mathbf{\ddot{x}=v_x}$ and $\dot{\Pi}_\mathrm{EPR}= v_\mathrm{EPR}$. This yields tracking error dynamics given as
\begin{align}
    \mathbf{\ddot{e}_x} &= \mathbf{v_x - \ddot{x}_{ref}}\,, \label{eq_FBL_e_x} \\[6pt]
    \dot{e}_\mathrm{EPR} &= v_\mathrm{EPR} - \dot{\Pi}_\mathrm{EPR,ref} \,, \label{eq_FBL_e_EPR} 
\end{align}
where the errors are defined as $\mathbf{e_x = x - x_{ref}}$ and $e_\mathrm{EPR} = \Pi_\mathrm{EPR} - \Pi_\mathrm{EPR,ref}$.

Relative degrees for spool speeds and EPR are 2 and 1, respectively. Based on this, the virtual inputs can be chosen as 
\begin{align}
    \mathbf{v_x} &= \mathbf{\ddot{x}_{ref} - K_{d,x}\,\dot{e}_x - K_{p,x}\,e_x}\,, \\[6pt]
    v_\mathrm{EPR} &= \dot{\Pi}_\mathrm{EPR,ref} - K_\mathrm{p,EPR}\,e_\mathrm{EPR}\,.
\end{align}

Inserting them into equations (\ref{eq_FBL_e_x}) and (\ref{eq_FBL_e_EPR}) yields
\begin{align}
    \mathbf{\ddot{e}_x} &= \mathbf{- K_{d,x}\,\dot{e}_x - K_{p,x}\,e_x}\,, \\[6pt]
    \dot{e}_\mathrm{EPR} &= - K_\mathrm{p,EPR}\,e_\mathrm{EPR}\,,
\end{align}
which are asymptotically stable systems for $\mathbf{K_{d,x} \succ 0}$, $\mathbf{K_{p,x} \succ 0}$, and ${K_\mathrm{p,EPR} > 0}$, and pole placement can be applied to tune the controller \cite{Bonfiglio2017, Slotine1991}.

\subsubsection{K-FBLC With Integrators}
To address steady-state offsets caused by modeling errors and varying operating conditions, an integrator-augmented version of the K-FBLC was also considered and evaluated. For spool-speed outputs, the virtual input was modified as
\begin{align}
    \mathbf{v_x} &= \mathbf{\ddot{x}_{ref}} - \mathbf{K_{d,x}\,\dot{e}_x} - \mathbf{K_{p,x}\,e_x} - \mathbf{K_{i,x}\,\eta_x}\,,  \\[6pt]
    \mathbf{\eta_x} &= \int_0^t \mathbf{e_x}(\tau) \mathrm{d}\tau\,.
\end{align}

For the EPR output, the virtual input was modified as
\begin{align}
    v_{\mathrm{EPR}} &= \dot{\Pi}_{\mathrm{EPR,ref}} - K_{p,\mathrm{EPR}}\,e_{\mathrm{EPR}} - K_{i,\mathrm{EPR}}\,\eta_{\mathrm{EPR}}\,, \\[6pt] \eta_{\mathrm{EPR}} &= \int_0^t e_{\mathrm{EPR}}(\tau) \mathrm{d}\tau\,.
\end{align}

\section{Results} \label{sec6}
The results are presented for two control strategies, demonstrating the flexibility of the Koopman model across different output selections. Validation is also performed for varying flight conditions.

\subsection{Target Engine Model Validation}
To ensure that the developed CLM is representative of a real GTE, it was validated against the GasTurb 15 commercial simulation software, which is the benchmark for engine modeling \cite{GasTurb}.

\begin{figure}[!ht] 
    \centering 
    \begin{subfigure}{0.45\textwidth}
        \centering
        \includegraphics[width=0.95\linewidth]{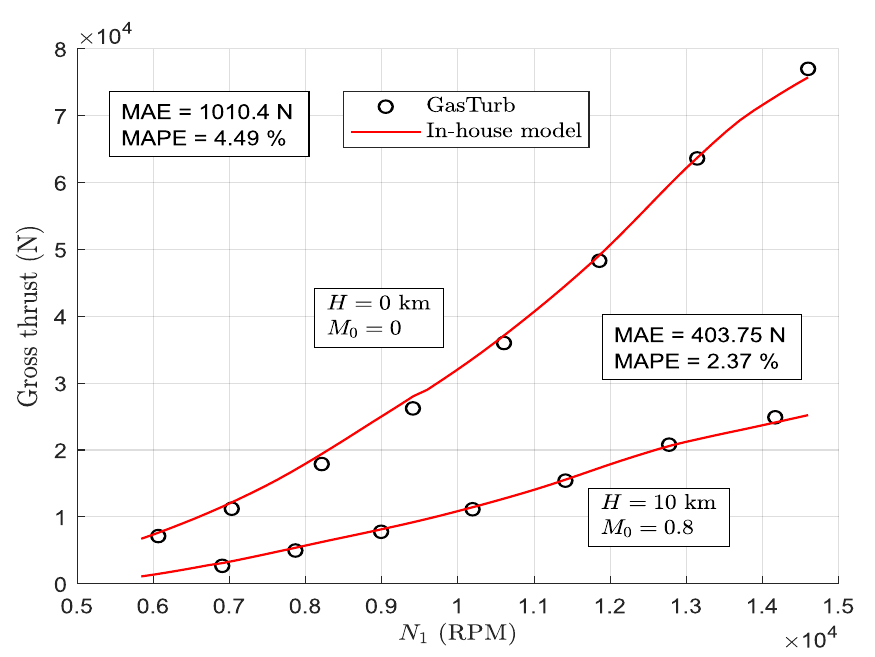}
        \caption{}
        \label{valid_SS_F}
    \end{subfigure}
    \begin{subfigure}{0.45\textwidth}
        \centering
        \includegraphics[width=1\linewidth]{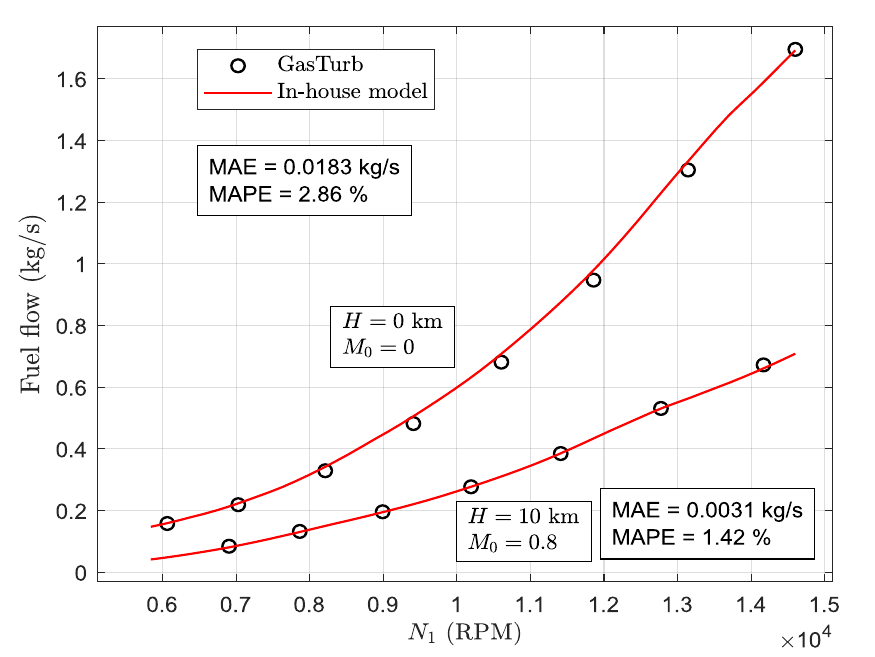}
        \caption{}
        \label{valid_SS_Wf}
    \end{subfigure}
    \caption{(a) Thrust and (b) fuel flow steady-state validation against the GasTurb 15 simulation software in $H = 0$ km, $M_0 = 0$ and $H = 10$ km, $M_0 = 0.8$.}
    \label{valid_SS}
\end{figure}

\begin{figure}[!ht]
    \centering
    \includegraphics[width=0.95\linewidth]{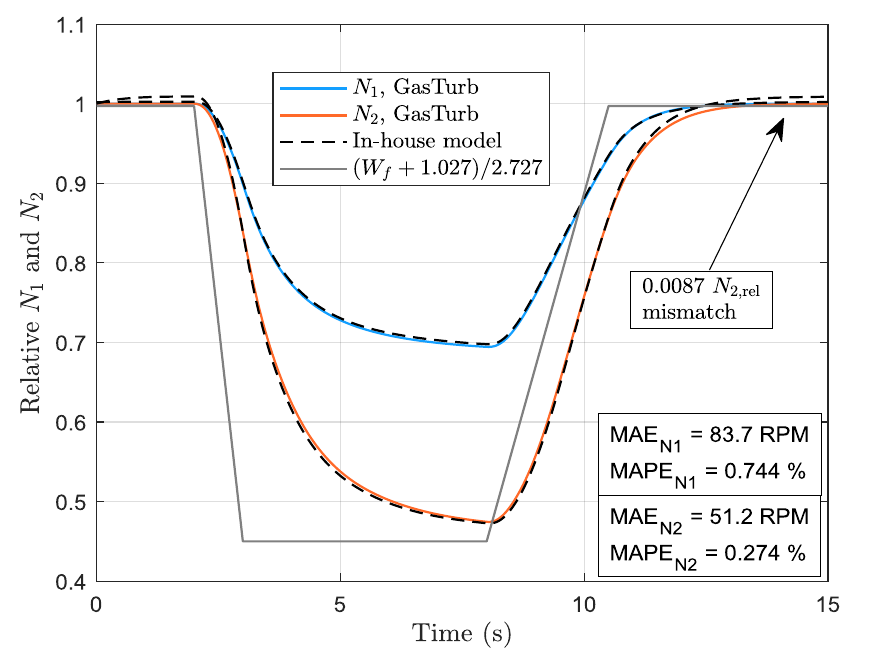}
    \caption{Transient performance validation against the GasTurb 15 simulation software for the sea-level conditions. The fuel flow is rescaled for better visualization.}
    \label{valid_TR}
\end{figure}

The model was validated for both steady-state and transient computations. In the steady state, validation was performed for sea-level and flight conditions. The results are depicted in Fig. \ref{valid_SS} and Fig. \ref{valid_TR}, and a quantitative summary is provided in Table \ref{tab_valid_SS} and Table \ref{tab_valid_TR}. The MAE and percentage MAE (MAPE) were selected. The results indicate that the in-house MATLAB model accurately represents the target engine's behavior.

\begin{table}[!ht]
\caption{MAE and MAPE values for the steady-state validation.}
\label{tab_valid_SS}
\centering
\begin{tabular*}{\columnwidth}{@{\extracolsep\fill}l|cc}
\toprule
\textbf{Steady-state} & $H = 0$ km, $M_0 = 0$ & $H = 10$ km, $M_0 = 0.8$ \\
\midrule
$\mathrm{MAE_{F}}$ & 1010.4 N & 403.75 N \\
$\mathrm{MAPE_{F}}$ & 4.49 $\%$ & 2.37 $\%$ \\
$\mathrm{MAE_{WF}}$ & 0.0183 kg/s & 0.0031 kg/s \\
$\mathrm{MAPE_{WF}}$ & 2.86 $\%$ & 1.42 $\%$ \\
\bottomrule
\end{tabular*}
\end{table}

\begin{table}[!ht]
\caption{MAE and MAPE values for the transient validation.}
\label{tab_valid_TR}
\centering
\begin{tabular*}{\columnwidth}{@{\extracolsep\fill}l|c}
\toprule
\textbf{Transient} & $H = 0$ km, $M_0 = 0$ \\
\midrule
$\mathrm{MAE_{N1}}$ & 83.7 RPM \\
$\mathrm{MAPE_{N1}}$ & 0.744 $\%$ \\
$\mathrm{MAE_{N2}}$ & 51.2 RPM \\
$\mathrm{MAPE_{N2}}$ & 0.274 $\%$ \\
\bottomrule
\end{tabular*}
\end{table}

\subsection{Identification of the Koopman Model}

\subsubsection{Training Dataset and Normalization}
The training dataset was obtained via closed-loop simulation of the dynamics, including quasi-amplitude-modulated pseudo-random 
binary sequences and sinusoidal changes in the setpoint. Gaussian white noise with a standard deviation of 0.25 $\%$ of the corresponding nominal values was added to the dataset, simulating the presence of expected measurement noise. To approximately exclude the explicit dependence of the dynamics on the flight conditions, the corrected parameters were utilized \cite{GrasevSpringer2026}, and the data were normalized as follows:
\begin{align}
    N_{\mathrm{1,norm}} &= \frac{N_\mathrm{1,corr} - 6000}{15800-6000}\,,\\[3pt] N_{\mathrm{2,norm}} &= \frac{N_\mathrm{2,corr} - 15000}{24000 - 15000}\,, \\[3pt]
    W_{\mathrm{f,norm}} &= \frac{W_\mathrm{f,corr}}{2.3}\,, \\[3pt]
    A_{\mathrm{n,norm}} &= \frac{A_\mathrm{n} - 0.19}{0.235-0.19}\,, \\[3pt]
    \Pi_\mathrm{EPR,norm} &= \frac{\Pi_\mathrm{EPR} - 1}{3.5}\,,
\end{align}
where the values are based on minimum and maximum values of the corresponding quantities, and corrected parameters are given as
\begin{align}
    N_\mathrm{1,corr} &= N_1\sqrt{\frac{288}{T_\mathrm{1t}}}\,, \\[6pt]
    N_\mathrm{2,corr} &= N_2\sqrt{\frac{288}{T_\mathrm{1t}}}\,, \\[6pt]
    W_\mathrm{f,corr} &= W_\mathrm{f}\frac{101325}{p_\mathrm{1t}}\sqrt{\frac{288}{T_\mathrm{1t}}}\,.
\end{align}

\begin{figure}[!ht]
    \centering
    \includegraphics[width=1\linewidth]{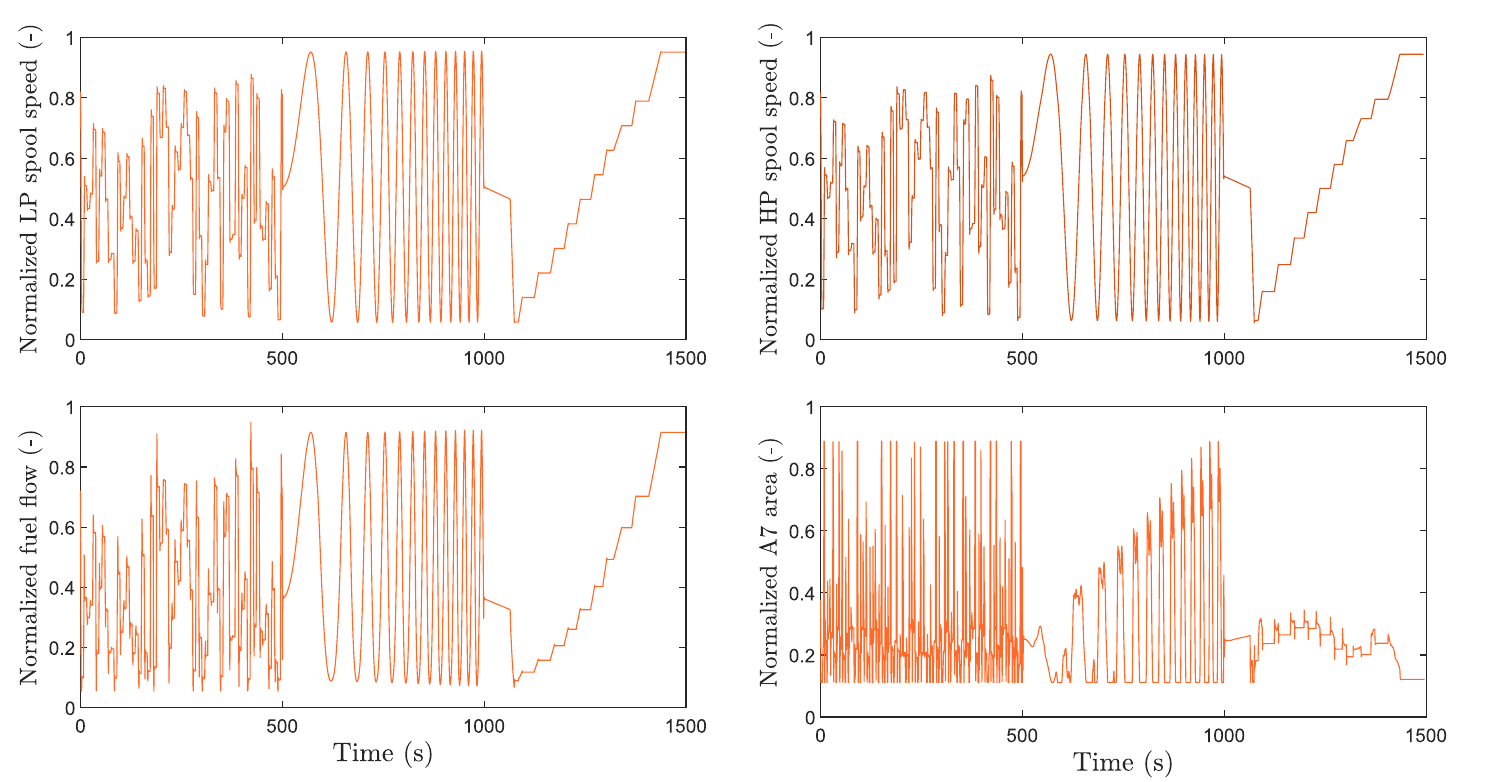}
    \caption{Dataset used for identification of the 2-spool GTE (without noise for clarity).}
    \label{2spool_dataset}
\end{figure}

It should be noted that, despite the data being generated by a closed-loop simulation, the inputs used for identification were the fuel flow and the nozzle area applied to the engine, yielding a dataset covering the behavior of the engine only. The dataset without noise is shown in Fig. \ref{2spool_dataset}.

\subsubsection{Metaheuristic EDMD}
The parameter vector in (\ref{eq_KOGA}) was first optimized using the particle swarm optimization (PSO) algorithm \cite{Kennedy1995-PSO, Eberhart2001-PSO} with 7 observables, a population size of 50, $\alpha = 10^{-4}$, and a stopping criterion of a maximum of 100 generations. Subsequently, the solution of PSO was utilized to initialize the Nelder-Mead simplex algorithm \cite{Nelder1965}, employed to refine the model parameters locally. The maximum number of generations was set to 1000. 

Initially, only a small segment of the dataset, with the time ranging from 300 to 700 seconds, covering the steps and sinusoidal, was utilized to warm-start the parameters. Subsequently, the solution was refined using the rest of the dataset.

To evaluate the effect of the selection of observables on the prediction, three functions were compared, namely the logistic function (LF), the Gaussian radial basis function (GAU RBF), and the inverse quadratic (IQ) RBF. The corresponding equations are
\begin{equation}  \label{eqRBF}
\begin{aligned}
    \mathrm{LF}(\textbf{x}) &= \frac{1}{1+\mathrm{exp}(\varepsilon_1 x_1 + \varepsilon_2 x_2 + b)}\,, \\[3pt]
    \mathrm{GAU}(\textbf{x}) &= \exp{\big(-\epsilon \|\mathbf{x - x_c}\|_2^2 \big)}\,, \\[3pt]
    \mathrm{IQ}(\textbf{x}) &= \frac{1}{1 + \epsilon \|\mathbf{x - x_c}\|_2^2}\,, 
\end{aligned}
\end{equation}
where $\mathbf{x_c}$ is a vector of collocation point coordinates, $\varepsilon_1$, and $\varepsilon_2$ are the shaping parameters, and $b$ is the LF bias term.

Besides the MAE criterion, the MAPE was also evaluated as
\begin{align}
    \mathrm{MAPE} = \frac{100}{N_t} \sum_{k = 1}^{N_t} \left|\frac{\mathbf{y_k - \hat{y}_k}}{\mathbf{y_k}}\right|\,.
\end{align}

The quantitative comparison is in Table \ref{tab_KOGA_comp}. The IQ observables were selected for the Koopman model, as they achieved an accurate prediction of both spool speeds and EPR.
\begin{table}[!ht] 
\caption{Comparison of the observable functions for MH-EDMD.}\label{tab_KOGA_comp}
\begin{tabular*}{\columnwidth}{@{\extracolsep\fill}l|ccc}
\toprule
Function & LF & GAU & IQ \\
\midrule
$\mathrm{MAE_{N1}}$ (RPM) & 31.92  & 19.14  & 20.3   \\[2pt]
$\mathrm{MAE_{N2}}$ (RPM) & 30.58  & 17.05  & 15.17  \\[2pt]
$\mathrm{MAE_{EPR}}$      & 0.0182 & 0.0116 & 0.0114 \\[2pt]
\midrule
$\mathrm{MAPE_{N1}}$ $\%$  & 1.52 & 0.656 & 0.68 \\[2pt]
$\mathrm{MAPE_{N2}}$ $\%$  & 1.76 & 0.55  & 0.6 \\[2pt]
$\mathrm{MAPE_{EPR}}$ $\%$ & 2.2  & 1.21  & 1.12 \\[2pt]
\bottomrule
\end{tabular*}
\end{table}

The Koopman system was subsequently converted to the eigenfunction form via eigen-decomposition of the $\mathbf{A}$ matrix and (\ref{eig_sys_conv}). Since $\mathbf{\Lambda}$ and $\mathbf{V}$ are generally matrices of complex numbers, the canonical transform was employed to obtain a real-valued system \cite{Surana2020}.

\begin{figure}[!ht]
    \centering
    \includegraphics[width=1\linewidth]{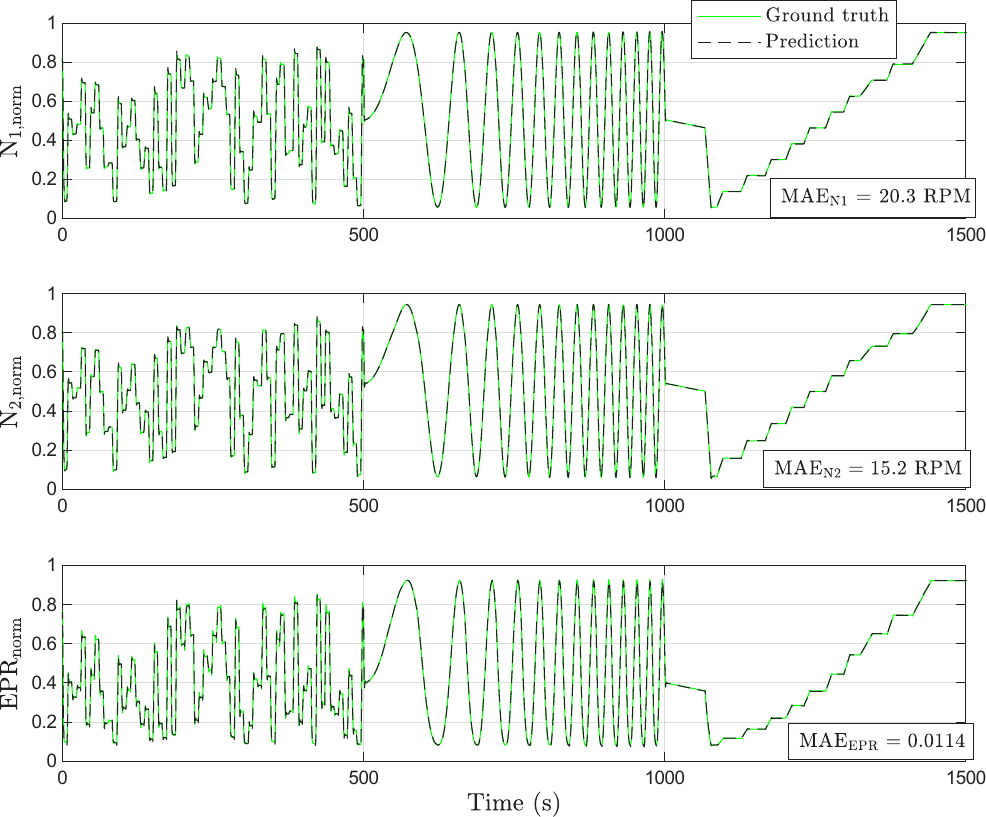}
    \caption{Prediction results of the identified Koopman system after numerical integration.}
    \label{KOGA_sys_prediction}
\end{figure}

The results for the IQ RBF are depicted in Fig. \ref{KOGA_sys_prediction}. It can be concluded that the obtained Koopman system provides highly accurate predictions of GTE behavior.

\subsubsection{Comparison to EDMDc with LTI Dynamics}
To compare the MH-EDMD for a time-varying system with the classical EDMDc with LTI input dynamics, the EDMDc was also performed using polynomials, Gaussian RBF, and IQ RBF augmented only with the inputs. The regularization parameter $\alpha$, number of observables $n_\Psi$, and the shaping parameters $\varepsilon$ of RBFs were the manipulated variables of a grid search analysis, and the collocation points were evenly distributed in the range $[0,1] \times [0,1]$. 

The best result, balancing the system order and accuracy, was obtained for 25 Gaussian RBFs with shaping parameter $\varepsilon = 10$, and regularization parameter $\alpha = 2$. The comparison with the MH-EDMD is provided in Table \ref{tab_EDMDc_comp}. All the identified time-varying systems outperform LTI systems by a large margin.
\begin{table}[!ht] 
\caption{Comparison of the MH-EDMD with the EDMDc with Gaussian RBF.}
\label{tab_EDMDc_comp}
\centering
\begin{tabular*}{\columnwidth}{@{\extracolsep\fill}l|cc}
\toprule
Metric & EDMDc, GAU & MH-EDMD, IQ \\
\midrule
$\mathrm{MAE}_{N_1}$ (RPM) & 372.3 & 20.3 \\
$\mathrm{MAE}_{N_2}$ (RPM) & 433.7 & 15.17 \\
$\mathrm{MAE}_{\mathrm{EPR}}$ & 0.0523 & 0.0114 \\
\bottomrule
\end{tabular*}
\end{table}

\subsubsection{Corrected Spool Speed Derivatives}
The spool speed time derivatives, $\dot{N}_1$ and $\dot{N}_2$, can also be corrected. However, it is important to note that $d/dt(N_\mathrm{corr}) \neq \dot{N}_\mathrm{corr}$. The valid correction reads \cite{Volponi2020}
\begin{align}
    \begin{bmatrix}
        \dot{N}_\mathrm{1,corr} \\ \dot{N}_\mathrm{2,corr}
    \end{bmatrix} = \begin{bmatrix}
        \dot{N}_1 \\ \dot{N}_2
    \end{bmatrix}\left(\frac{101325}{p_\mathrm{1t}}\right)\,.
\end{align}

Therefore, as discussed in \cite{GrasevAccess2025}, the system (\ref{eig_sys_conv}) is multiplied by $p_\mathrm{1t}/101325$ for the prediction of spool speeds in varying flight conditions.

\subsection{$N_1-N_2$ Control}
The reference tracking was evaluated using a stair sequence with 5 small steps, full acceleration, and full deceleration. The results are shown for relative physical spool speeds computed as $N_\mathrm{1,rel} = N_{1}/14600$ and $N_\mathrm{2,rel} = N_{2}/22800$. For all subsequent scenarios, the measurement noise was modeled by Gaussian white noise with standard deviations of 30 RPM and 0.001 for spool speeds and EPR, respectively.

\subsubsection{Controller Tuning}
PI controllers are commonly used for GTE control. The benchmark PI controller was manually tuned to respond to a 1-second ramp command from idle to maximum thrust in 5 seconds with no overshoots of the spool speeds. Decoupled loops were considered, where the fuel flow affects mainly the high-pressure turbine inlet temperature and $N_2$ dynamics, and the nozzle area affects primarily the low-pressure turbine pressure ratio and $N_1$ dynamics. Thus, the control law is
\begin{equation}
    \begin{aligned}
        W_\mathrm{f}(t) = K_\mathrm{p2}e_\mathrm{N2}(t) + K_\mathrm{i2}\int_{0}^t e_\mathrm{N2}(\tau) \mathrm{d}\tau\,, \\[6pt]
        A_\mathrm{n}(t) = K_\mathrm{p1}e_\mathrm{N1}(t) + K_\mathrm{i1}\int_{0}^t e_\mathrm{N1}(\tau) \mathrm{d}\tau\,,
    \end{aligned}
\end{equation}
where $e_\mathrm{Ni} = N_\mathrm{i,ref} - N_i$, $i = 1,2$.

The resulting gains were $K_\mathrm{p1} = 6.85 \times 10^{-4}$, $K_\mathrm{i1} = 0.0021$, $K_\mathrm{p2} = 4.4 \times 10^{-4}$, and $K_\mathrm{i2} = 0.0011$. 

The K-FBLC gain matrices were tuned as $\mathbf{K_{p,x}}= \mathrm{diag}(90,120)$ and $\mathbf{K_{d,x}}= \mathrm{diag}(15,20)$.

The K-FBLC-I gains were tuned as $\mathbf{K_{p,x}} = \mathrm{diag}(120, 150)$, $\mathbf{K_{d,x}} = \mathrm{diag}(10,20)$, and $\mathbf{K_{i,x}} = [3\,\,2;\,\,2\,\,5]$. 

The MPC controller was tuned using $\mathbf{Q_y} = 30\mathbf{I}$ and $\mathbf{R}=0.01\mathbf{I}$. After extensive analysis, the prediction and control horizons were set to $n_p = 30$ and $n_c = 5$, respectively, to balance computational time, response speed, and oscillations. Prediction horizons below 20 led to a significant degradation in control performance. The augmented Kalman filter with DO was tuned using $\mathbf{Q_\Phi} = 10\mathbf{I}$, $\mathbf{Q_d} = 50\mathbf{I}$, and $\mathbf{R_{o} = I}$. The input increment constraints $\mathbf{\Delta u} \in [-0.25,0.25]$. The size of the problem was thus 10 decision variables, 20 input increment inequalities, and 20 input absolute value inequalities. The solver used was the native MATLAB \texttt{quadprog} with the active set method initialized at $\mathbf{\Delta U_k = 0}$ (steady state).

\subsubsection{Sea-Level Conditions}
\begin{figure}[!ht]
    \centering
    \includegraphics[width=1\linewidth]{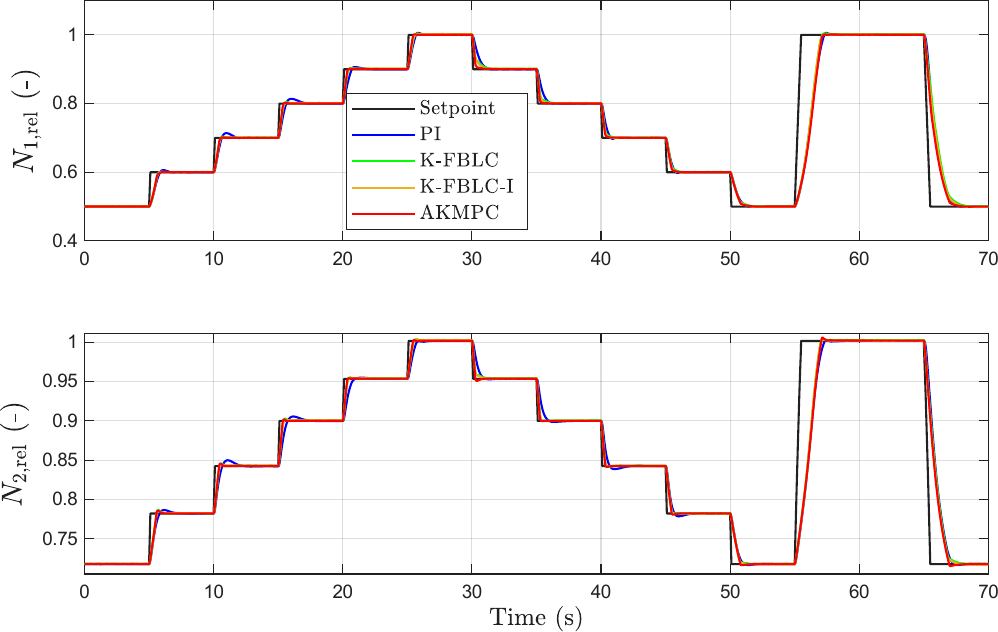}
    \caption{Spool speed response in sea-level conditions.}
    \label{N1_N2_SL_Spool_Speeds}
\end{figure}

\begin{figure}[!ht] 
    \centering 
    \begin{subfigure}{0.45\textwidth}
        \centering
        \includegraphics[width=0.9\linewidth]{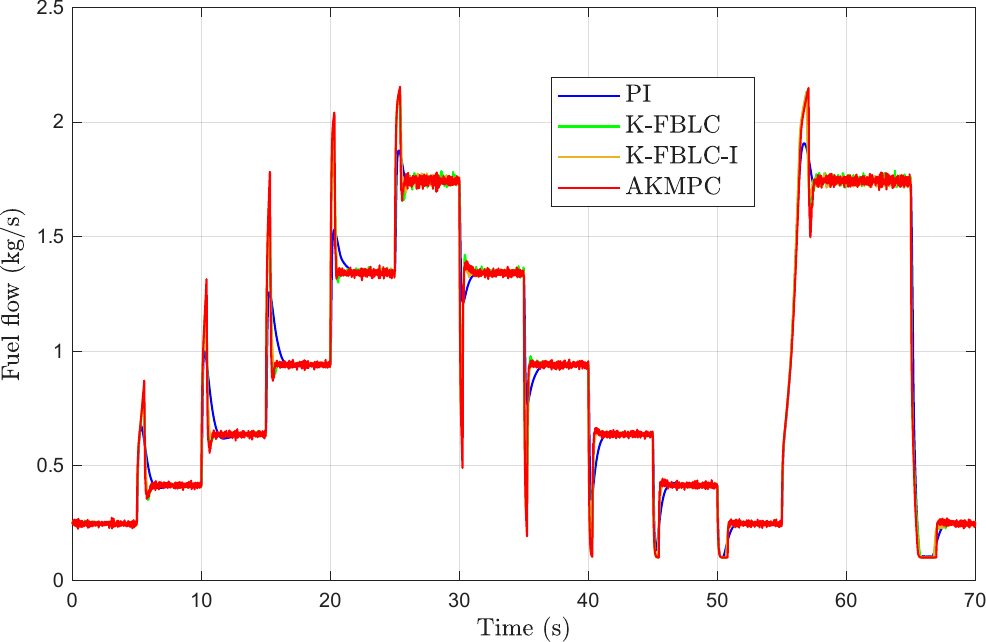}
        \caption{}
        \label{N1_N2_SL_Wf}
    \end{subfigure}
    \begin{subfigure}{0.45\textwidth}
        \centering
        \includegraphics[width=0.9\linewidth]{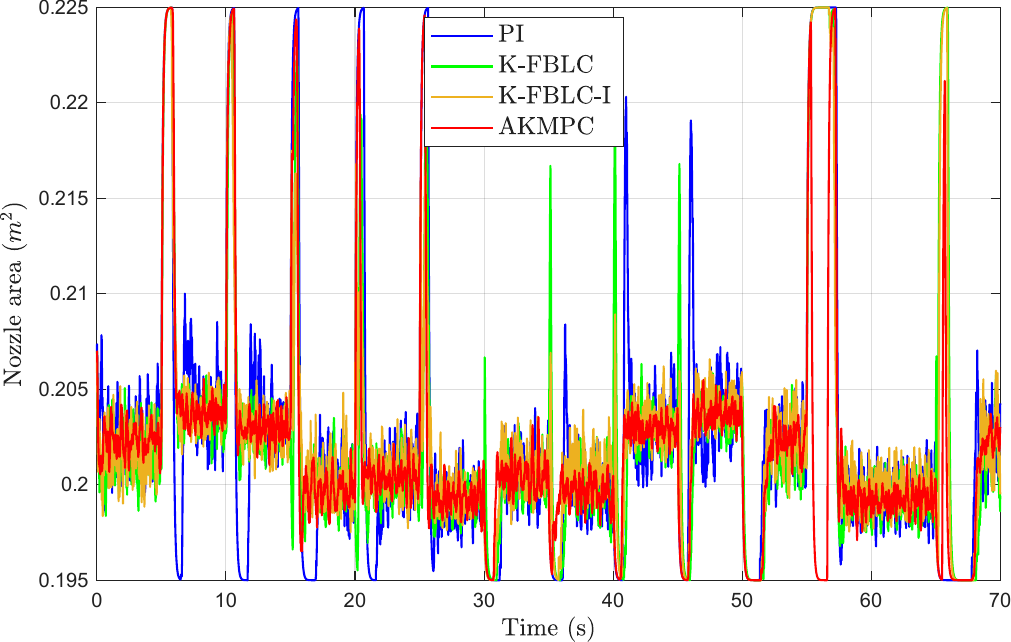}
        \caption{}
        \label{N1_N2_SL_An}
    \end{subfigure}
    \caption{(a) Fuel flow and (b) nozzle area comparison for $N_1-N_2$ in sea-level conditions.}
    \label{N1_N2_SL_Inputs}
\end{figure}

The comparison of PI, K-FBLC, K-FBLC-I, and AKMPC in sea-level conditions is depicted in Fig. \ref{N1_N2_SL_Spool_Speeds}, and the inputs are shown in Fig. \ref{N1_N2_SL_Inputs}. 

The AKMPC provides the best tracking performance in terms of IAE, overshoots, and noise in the control inputs. The settling times for K-FBLC and K-FBLC-I are on par with those of the AKMPC during acceleration, but slightly longer during deceleration in the high-RPM region. This also holds for the PI with the additional cost of larger overshoots. This reflects that a classical linear PI controller, unlike the other two methods, cannot adequately capture the nonlinear dynamics in the GTE's intermediate operating range. Note that during the initial stage of full acceleration and deceleration, the safety limits constrain the inputs, yielding a similar performance for all controllers. However, in the terminal stage, where the outputs reach the setpoint, the controller performance is distinguishable, and the AKMPC adheres more closely to the setpoint thanks to the terminal cost function, utilizing the full range of $W_\mathrm{f}$ limits with less abrupt $A_\mathrm{n}$ changes. In addition, the AKMPC increases $A_\mathrm{n}$ immediately after a positive change in the $N_1$ setpoint, and vice versa, thereby enhancing response speed.

Table \ref{tab_SL_IAE} shows the integral of absolute error (IAE) criterion used for quantifying control performance. It is approximated as
\begin{align} \label{eq_IAE}
    \mathrm{IAE} &= \sum_{i=1}^{t_\mathrm{end}/\Delta t} { | \mathbf{y_{set,i} - y_i} | } \,.
\end{align}

\begin{figure}[!ht]
    \centering
    \includegraphics[width=0.9\linewidth]{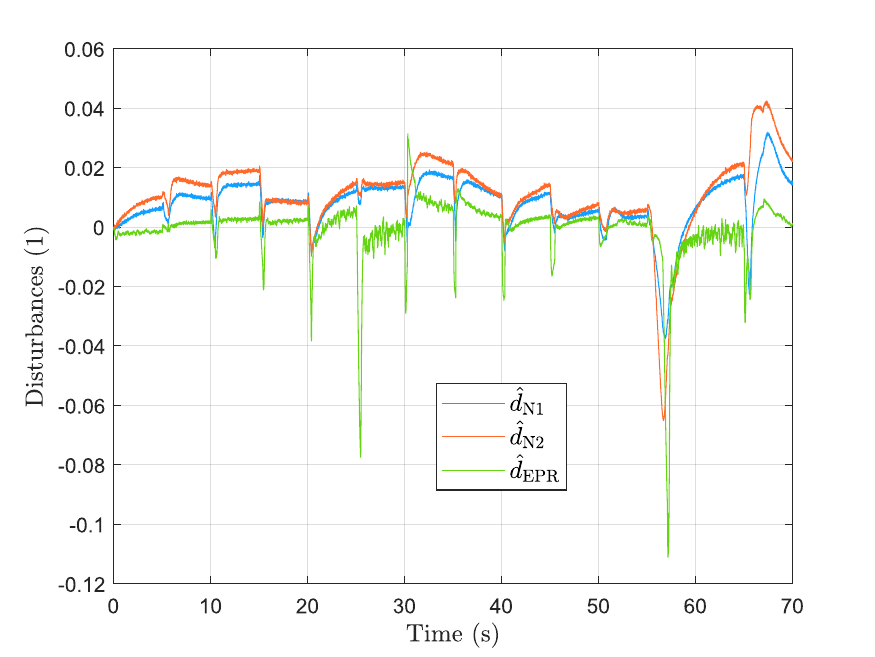}
    \caption{Estimated disturbances for the $N_1-N_2$ case in sea-level conditions.}
    \label{N1_N2_SL_Dist}
\end{figure}

Estimated disturbances are shown in Fig. \ref{N1_N2_SL_Dist}. The largest corrections were applied to the EPR, with a maximum of 0.11.

\begin{table}[!ht] 
    \caption{The IAE criterion values for the controllers in sea-level conditions.}\label{tab_SL_IAE}
    \begin{tabular*}{\columnwidth}{@{\extracolsep\fill}l|cc}
    \toprule
    Controller & $\mathrm{IAE}\ N_1$ & $\mathrm{IAE}\ N_2$ \\[2pt]
    Unit & $\mathrm{RPM\,s}$ & $\mathrm{RPM\,s}$ \\
    \midrule
    PI & 17381 & 14600 \\[2pt]
    K-FBLC  & 16763 & 13526 \\[2pt]
    K-FBLC-I & 15740 & 12660 \\[2pt]
    AKMPC & 14869 & 11744 \\
    \bottomrule
    \end{tabular*}
\end{table}

\subsubsection{Varying Flight Conditions}
\begin{figure}[!ht] 
    \centering 
    \begin{subfigure}{0.45\textwidth}
        \centering
        \includegraphics[width=0.9\linewidth]{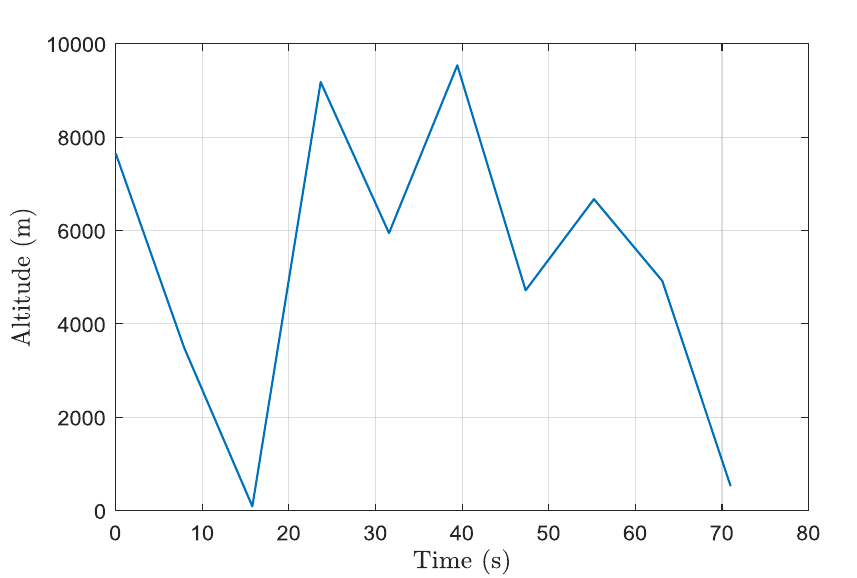}
        \caption{}
        \label{2spool_VFC_Altitude}
    \end{subfigure}
    \begin{subfigure}{0.45\textwidth}
        \centering
        \includegraphics[width=0.9\linewidth]{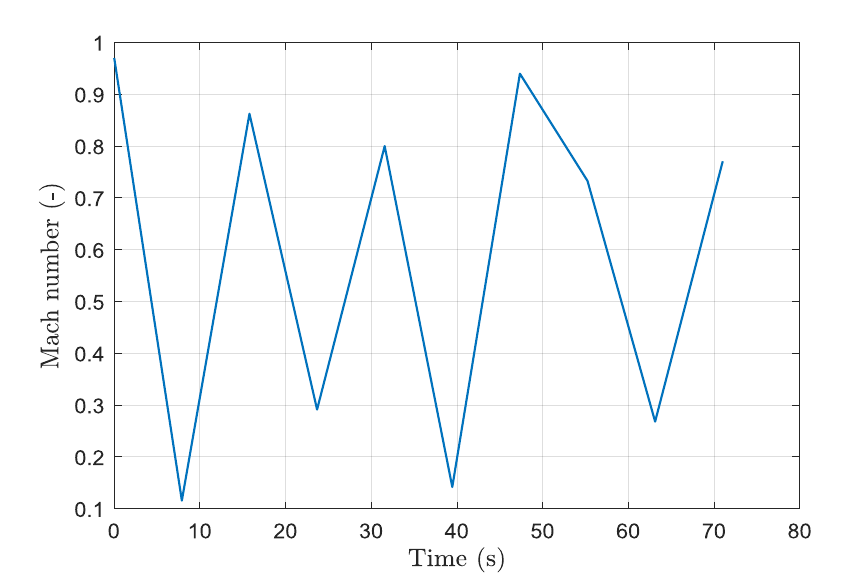}
        \caption{}
        \label{2spool_VFC_MachNumber}
    \end{subfigure}
    \caption{(a) Altitude and (b) Mach number of flight for the evaluation.}
    \label{2spool_VFC_Alt_Mach}
\end{figure}

To validate the effectiveness of parameter corrections, the control performance was also evaluated in varying flight conditions. Randomly generated altitude and Mach number profiles are shown in Fig. \ref{2spool_VFC_Alt_Mach}. These changes in flight conditions are purposely exaggerated and are not encountered during normal operation.

\begin{figure}
    \centering
    \includegraphics[width=1\linewidth]{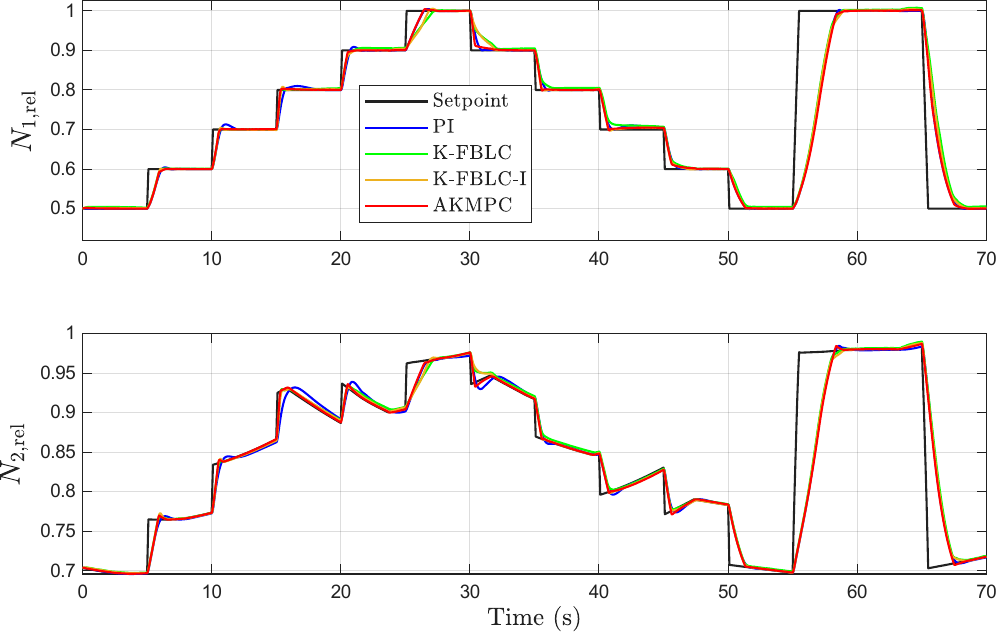}
    \caption{Spool speed response in varying flight conditions.}
    \label{N1_N2_VFC_Spool_Speeds}
\end{figure}

The qualitative comparison of results for the controllers is provided in Fig. \ref{N1_N2_VFC_Spool_Speeds} and IAE values are summarized in Table \ref{tab_VFC_IAE}. The AKMPC again exhibits the best performance in terms of IAE and adherence to the setpoint. Interestingly, the K-FBLC exhibits the highest IAE for $N_1$, with a slower response, particularly in the high-RPM region at higher altitude and Mach number. However, qualitatively, the K-FBLC and K-FBLC-I outperform the PI controller in terms of overshoots. The integrators successfully mitigate the steady-state offset, improving the performance of K-FBLC.

\begin{figure}[!ht] 
    \centering    
        \centering
        \includegraphics[width=1\linewidth]{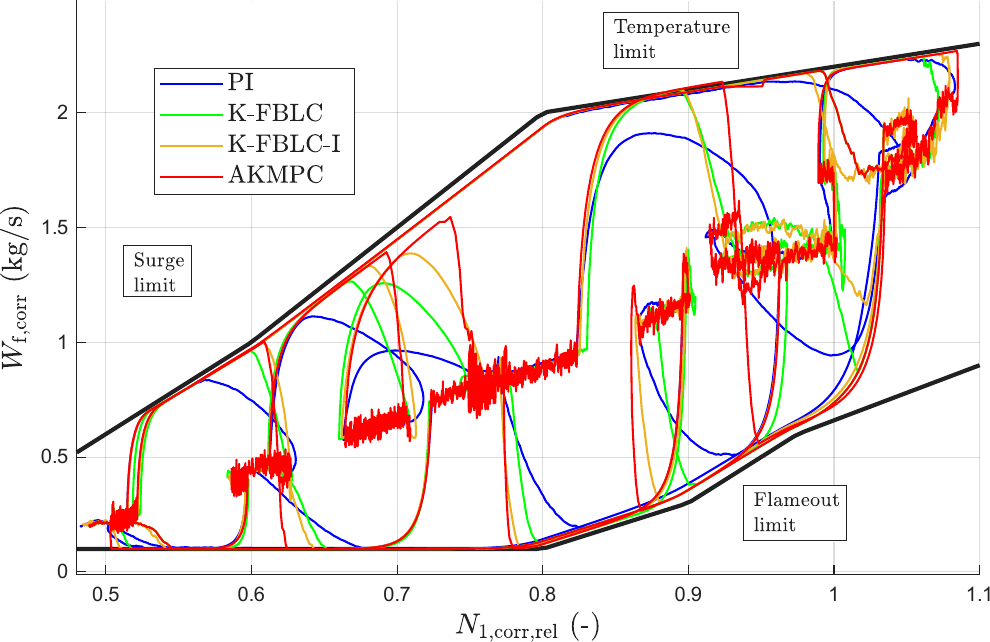}
        \caption{Comparison of fuel flow with limits in varying flight conditions.}
        \label{N1_N2_VFC_Wf}
\end{figure}

The fuel flow command comparison in Fig. \ref{N1_N2_VFC_Wf} demonstrates that all controllers respected the limits of corrected fuel flow. The AKMPC fully exploited the range by maintaining the fuel flow at its limits until shortly before reaching the setpoint, resulting in a faster transient response.

It should be noted that achieving fast tracking with minimum overshoot is of paramount importance for GTE control. Therefore, it was concluded that the AKMPC achieved the best performance, taking into account not only the IAE but also overshoots and settling times. 

\begin{figure}[!ht] 
    \centering 
    \begin{subfigure}{0.45\textwidth}
        \centering
        \includegraphics[width=1\linewidth]{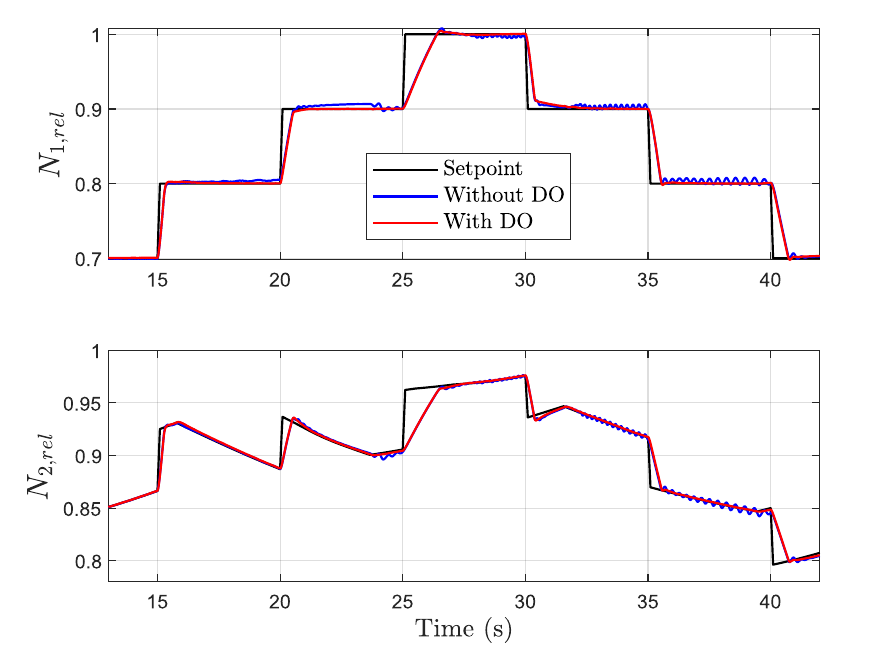}
        \caption{}
        \label{N1_N2_VFC_DOA_Spool_Speeds}
    \end{subfigure}
    \begin{subfigure}{0.45\textwidth}
        \centering
        \includegraphics[width=0.8\linewidth]{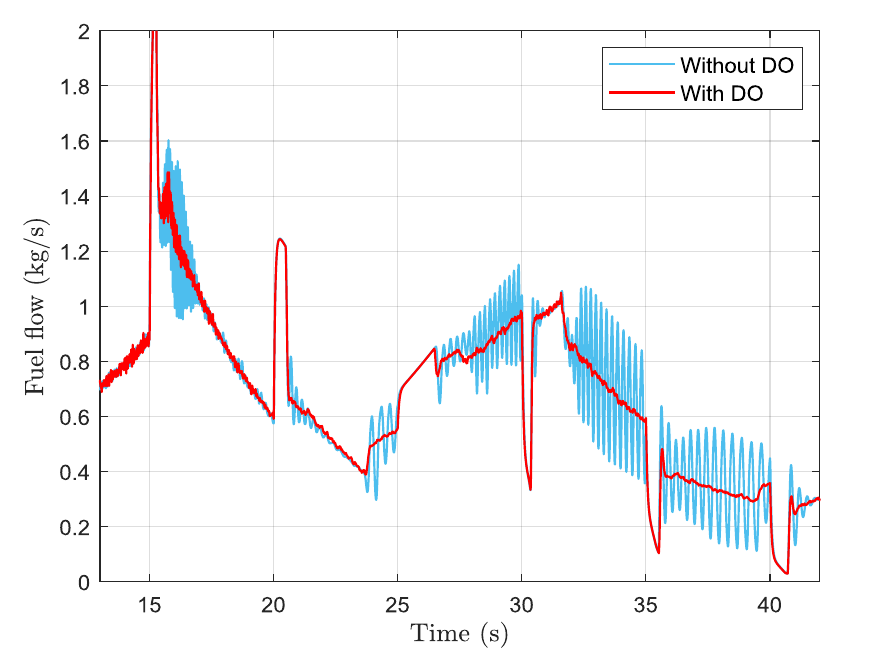}
        \caption{}
        \label{N1_N2_VFC_DOA_Fuel_Flow}
    \end{subfigure}
    \begin{subfigure}{0.45\textwidth}
        \centering
        \includegraphics[width=0.8\linewidth]{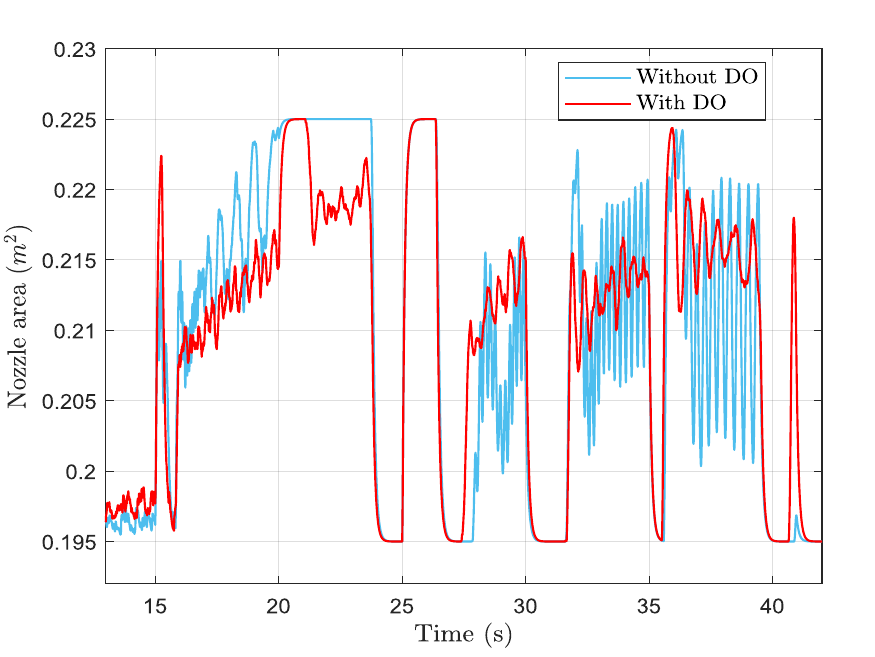}
        \caption{}
        \label{N1_N2_VFC_DOA_Nozzle_Area}
    \end{subfigure}
    \caption{(a) Spool speeds, (b) fuel flow, and (c) nozzle area comparison for AKMPC with and without the DO in varying flight conditions.}
    \label{N1_N2_VFC_DOA}
\end{figure}

\begin{figure}[!ht]
    \centering
    \includegraphics[width=0.9\linewidth]{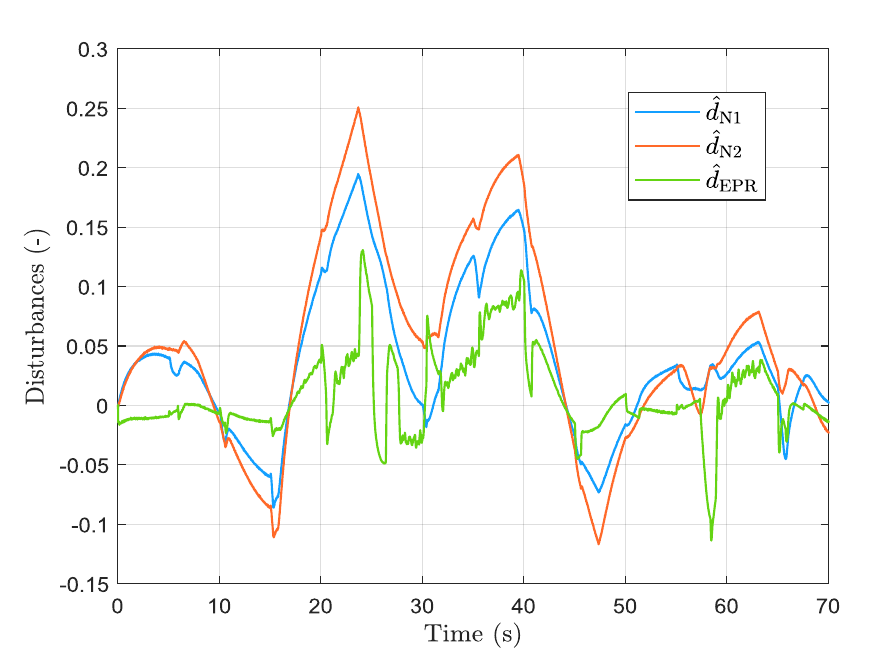}
    \caption{Estimated disturbances for the $N_1-N_2$ case in varying flight conditions.}
    \label{N1_N2_VFC_Dist}
\end{figure}

To show the effect of DO, AKMPC simulations were also performed with DO disabled. The results in Fig. \ref{N1_N2_VFC_DOA} indicate that the DO significantly improves performance, mitigating excessive oscillations caused by the model mismatch, as the engine parameter corrections apply effectively only in subsonic flight. The estimated disturbances are depicted in Fig. \ref{N1_N2_VFC_Dist}. The DO effectively accounts for the model mismatch and changes in flight conditions. The largest corrections were applied to $N_2$ with a maximum of 0.25.

The results indicate the applicability of both Koopman controllers for turbofan GTE control even in varying flight conditions.

\begin{table}[!ht] 
    \caption{The IAE criterion values for the controllers in varying flight conditions.}\label{tab_VFC_IAE}
    \begin{tabular*}{\columnwidth}{@{\extracolsep\fill}l|cc}
    \toprule
    Controller & $\mathrm{IAE}\ N_1$ & $\mathrm{IAE}\ N_2$ \\[2pt]
    Unit & $\mathrm{RPM\,s}$ & $\mathrm{RPM\,s}$ \\
    \midrule
    PI & 24632 & 22621  \\[2pt]
    K-FBLC  & 27971 & 21807  \\[2pt]
    K-FBLC-I & 25760 & 20710 \\[2pt]
    AKMPC & 23835 & 19548 \\
    \bottomrule
    \end{tabular*}
\end{table}

\subsubsection{Computational Time Analysis}
The per-step computational times were measured for the AKMPC and Koopman predictor modules of the main simulation code using the MATLAB \texttt{tic-toc} function. All computations were performed under controlled desktop conditions using a
laptop with the specifications summarized in Table \ref{tab_laptop_specs}. During the analysis, the only application running was MATLAB.
\begin{table}[!ht] 
    \caption{Hardware and software specifications.}\label{tab_laptop_specs}
    \begin{tabular*}{\columnwidth}{@{\extracolsep\fill}l|c}
    \toprule
    Item & Specification \\
    \midrule
    CPU & AMD Ryzen 7 8845HS 3.8 GHz \\[2pt]
    RAM & 32 GB DDR5  \\[2pt]    
    Operating System & Microsoft Windows 11 10.0.26200 \\[2pt]
    MATLAB Version  & R2025b Update 4 \\
    \bottomrule
    \end{tabular*}
\end{table}

The mean, standard deviation, and median of the per-step time were evaluated in the sea-level and varying flight conditions across multiple simulations. The times are listed in Table \ref{tab_perstep_times}. The mean and median per-step times were below 1 ms in all cases, indicating the computational tractability of the AKMPC approach.
\begin{table}[!ht] 
    \caption{The per-step computational times for the AKMPC and Koopman predictor.}\label{tab_perstep_times}
    \begin{tabular*}{\columnwidth}{@{\extracolsep\fill}l|ccc}
    \toprule
    Case & Mean (ms) & STD (ms) & Median (ms) \\
    \midrule
    SL & 0.94 & 0.27 & 0.86  \\[2pt]
    VFCs & 0.98 & 0.30 & 0.88  \\
    \bottomrule
    \end{tabular*}
\end{table}

\subsubsection{Effects of the Limiters}
\begin{figure}[!ht] 
    \centering 
    \begin{subfigure}{0.45\textwidth}
        \centering
        \includegraphics[width=0.9\linewidth]{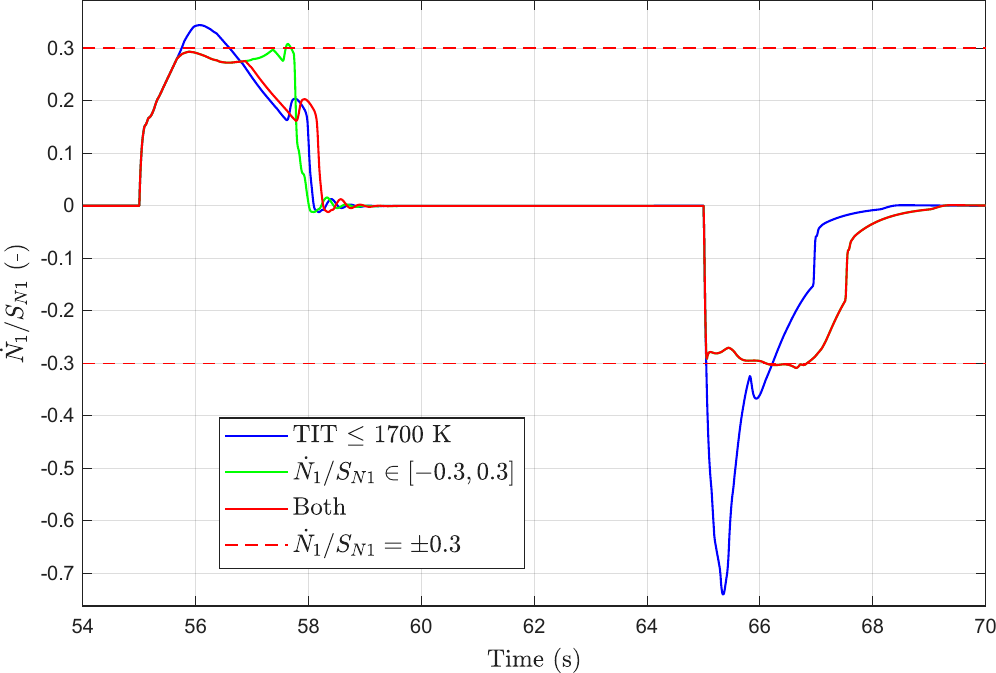}
        \caption{}
        \label{limiters_N1dot}
    \end{subfigure}
    \begin{subfigure}{0.45\textwidth}
        \centering
        \includegraphics[width=0.95\linewidth]{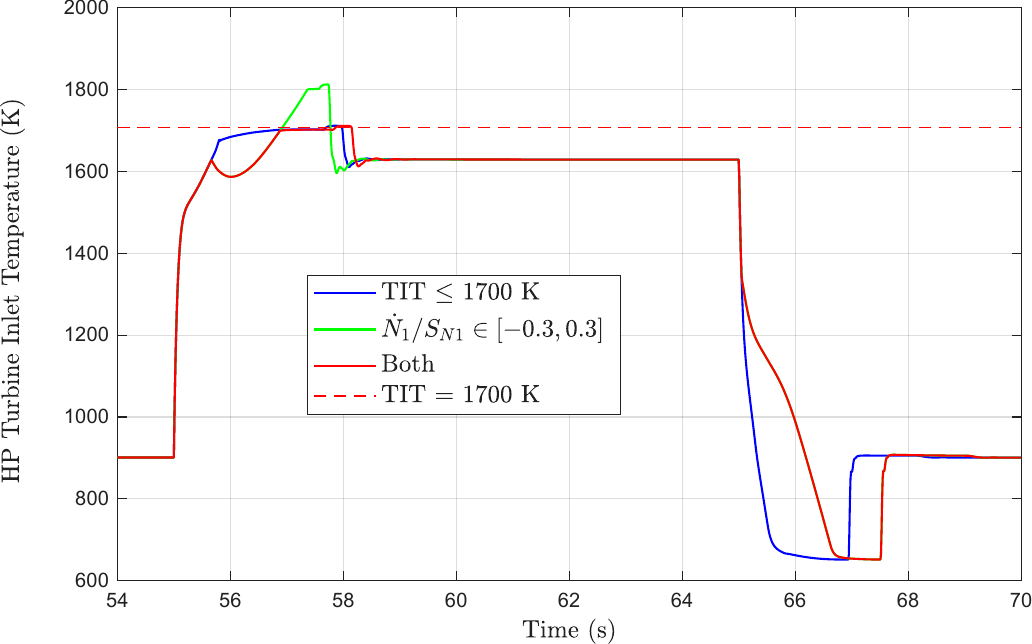}
        \caption{}
        \label{limiters_TIT}
    \end{subfigure}
    \caption{Effect of the TIT and $\dot{N}_1$ limiters on the limited outputs - (a) the $\dot{N}_1$ and (b) the TIT. The limiters successfully prevented the system from exceeding the set operational limits.}
    \label{limiters}
\end{figure}

In the previous simulations, only input constraints were imposed. These input bounds were selected such that the surge, over-temperature, and blowout limits were accounted for implicitly. To demonstrate the explicit output constraints from Section \ref{sec_cons}, additional simulations were performed with TIT and $\dot{N}_1$ constraints. The model prediction MAE values for TIT and $\dot{N}_1$ were 12.3 K and 16.8 RPM/s, respectively. The QP size increased with 15 new constraints, and the per-step time median was 1.25 ms. Figure \ref{limiters} shows their effect during the final large transient. The TIT limit was set to 1700 K, approximately 95 $\%$ of the maximum value implicitly covered by the input constraints. As shown, violation of the prescribed limits was successfully avoided, with the optimizer automatically restricting the control action according to the active constraints. 

\subsection{$\Pi_\mathrm{EPR}-N_1$ Control}
\begin{figure}[!ht]
    \centering
    \includegraphics[width=0.9\linewidth]{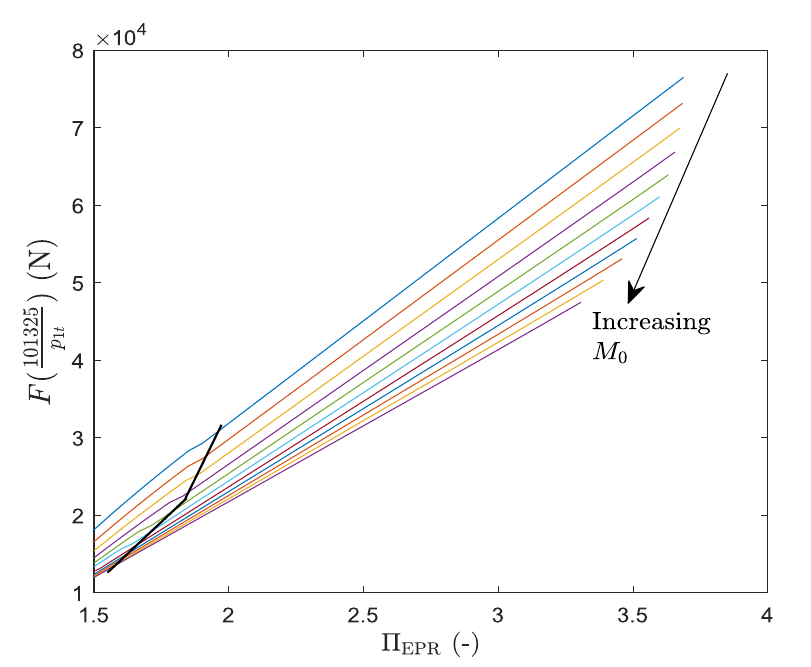}
    \caption{The relation between the corrected thrust and engine pressure ratio across varying Mach numbers.}
    \label{fig_F_vs_EPR}
\end{figure}

Control using EPR can improve indirect control of the thrust \cite{Jaw2009, Garg1989}, since there is an approximately linear relation between the corrected thrust and EPR across varying Mach numbers, as depicted in Fig. \ref{fig_F_vs_EPR}, compared to the relation to the corrected $N_1$ speed in Fig. \ref{valid_SS_F}. This particularly holds in the medium- to high-RPM range, where the engine operates most of the time. 

The commands for EPR and $N_1$ were now generated based on a commanded thrust profile with the same steps as in the previous case, ranging from 10000 N to 70000 N.

\subsubsection{Controller Tuning}
For this control strategy, only the K-FBLC and AKMPC controllers were compared. The K-FBLC gains were now $K_\mathrm{p,N1} = 100$, $K_\mathrm{d,N1} = 25$, and $K_\mathrm{p,EPR} = 40$.

The K-FBLC-I gains were tuned as $k_\mathrm{p,N1} = 50$, $k_\mathrm{d,N1} = 10$, $k_\mathrm{p,EPR} = 50$, and $\mathbf{K_i} = \mathrm{diag}(2,10)$.

The MPC controller was tuned using $\mathbf{Q_y} = \mathrm{diag}(20,100)$ and $\mathbf{R}=0.01\mathbf{I}$. The prediction and control horizons were kept $n_p = 30$ and $n_c = 5$, respectively. The augmented Kalman filter with DO was tuned using $\mathbf{Q_\Phi} = 10\mathbf{I}$, $\mathbf{Q_d} = 50\mathbf{I}$, and $\mathbf{R_{o} = I}$. The input increment constraints $\mathbf{\Delta u} \in [-0.25,0.25]$.

\subsubsection{Sea-Level Conditions}
\begin{figure}
    \centering
    \includegraphics[width=1\linewidth]{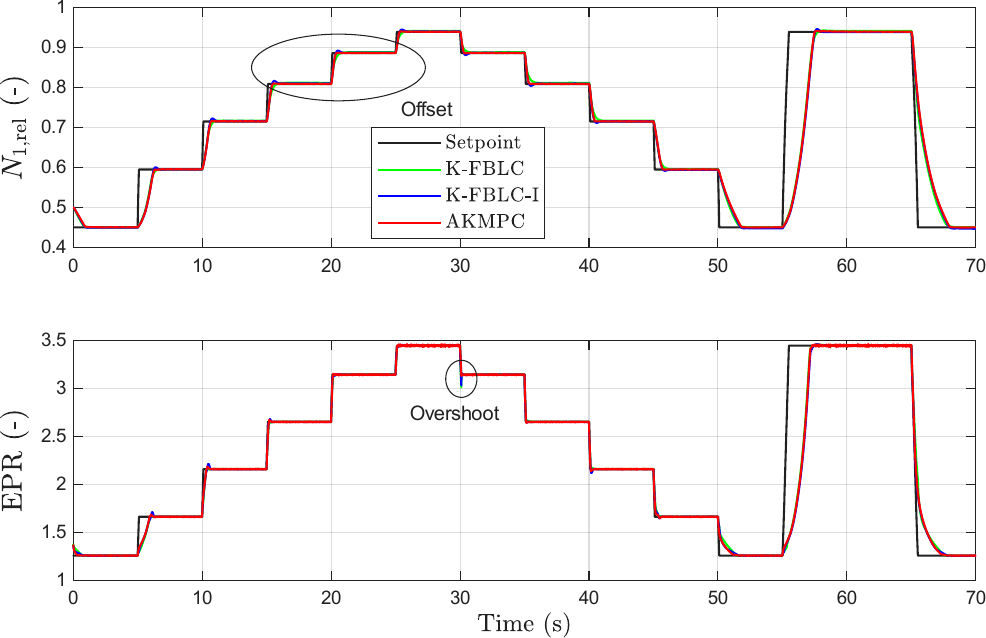}
    \caption{EPR and $N_1$ response in sea-level conditions. Black circles mark K-FBLC and K-FBLC-I overshoots.}
    \label{EPR_N1_SL_EPR}
\end{figure}

The comparison of the controllers in sea-level conditions is depicted in Fig. \ref{EPR_N1_SL_EPR}, and the IAE is summarized in Table \ref{tab_EPR_SL_IAE}. The performance is similar, with the AKMPC slightly outperforming the K-FBLC, which exhibits nonzero offset errors, and the K-FBLC-I, where the offset is mitigated by integrators, but multiple overshoots occur.

\begin{figure}[!ht] 
    \centering 
    \begin{subfigure}{0.45\textwidth}
        \centering
        \includegraphics[width=0.9\linewidth]{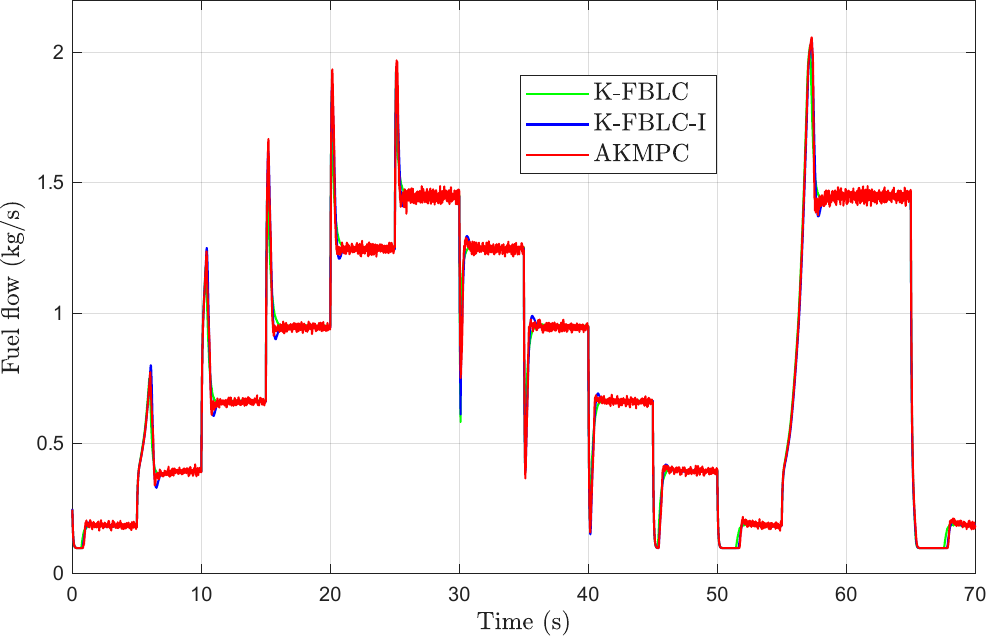}
        \caption{}
        \label{EPR_N1_SL_Wf}
    \end{subfigure}
    \begin{subfigure}{0.45\textwidth}
        \centering
        \includegraphics[width=0.9\linewidth]{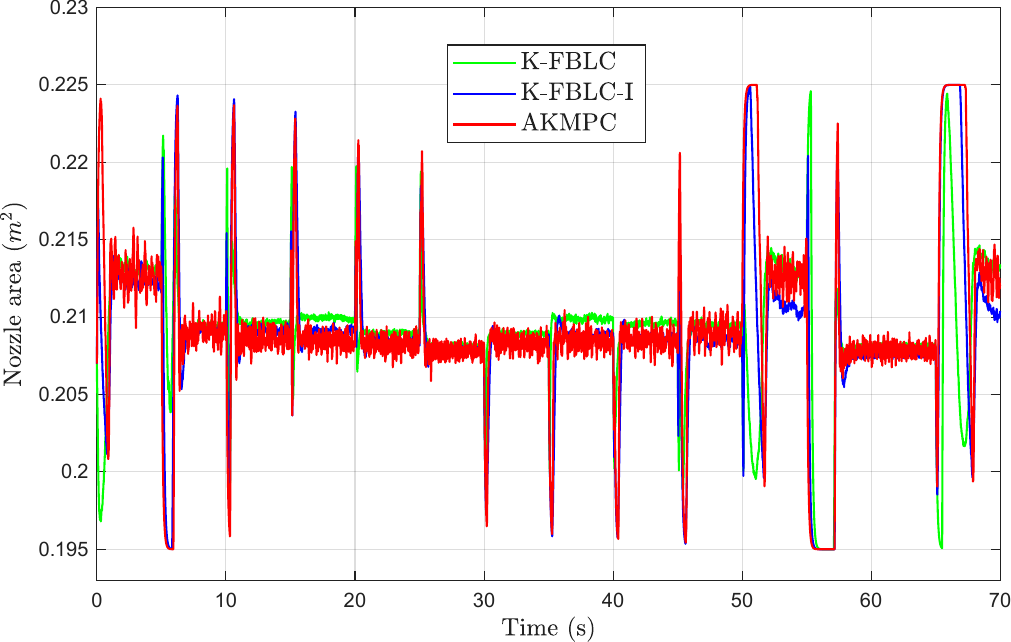}
        \caption{}
        \label{EPR_N1_SL_An}
    \end{subfigure}
    \caption{(a) Fuel flow and (b) nozzle area comparison for $\Pi_\mathrm{EPR}-N_1$ in sea-level conditions.}
    \label{EPR_N1_SL_Inputs}
\end{figure}

The inputs are shown in Fig. \ref{EPR_N1_SL_Inputs}. As can be seen, both controllers mostly increase $A_\mathrm{n}$ with a positive setpoint change, and vice versa, in the high-RPM region, with the AKMPC relying more on $W_\mathrm{f}$ in the low-RPM region. The inputs of AKMPC are noisier. 

\begin{figure}[!ht]
    \centering
    \includegraphics[width=0.9\linewidth]{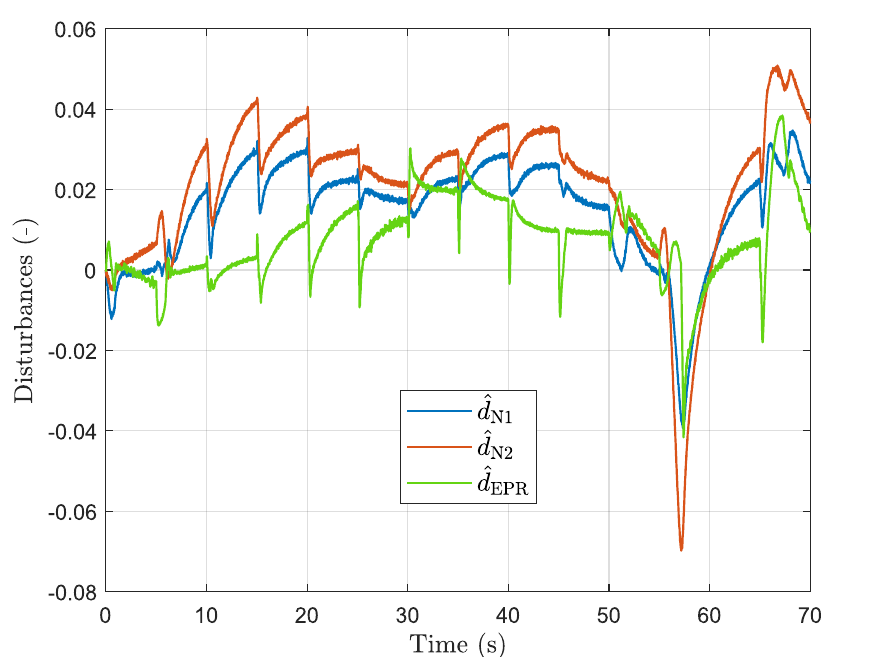}
    \caption{Estimated disturbances for the $\Pi_\mathrm{EPR}-N_1$ case in sea-level conditions.}
    \label{EPR_N1_SL_Dist}
\end{figure}

The estimated disturbances are shown in Fig. \ref{EPR_N1_SL_Dist}. Compared to Fig. \ref{N1_N2_SL_Dist}, the corrections were smaller, with a maximum of 0.07.

\begin{figure}[!ht]
    \centering
    \includegraphics[width=1\linewidth]{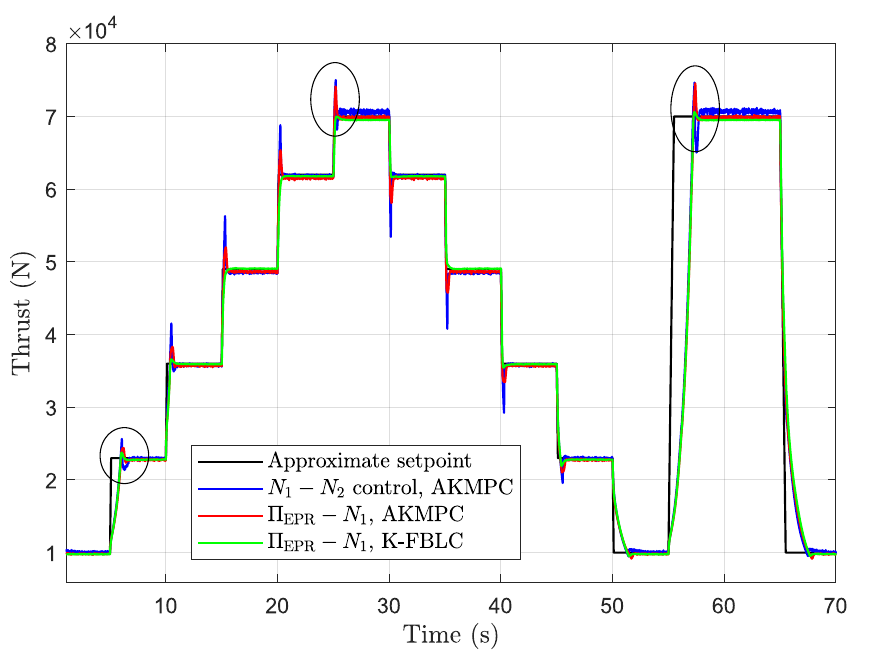}
    \caption{Thrust comparison for the two control strategies. Black circles mark regions with excessive oscillations for the $N_1-N_2$ controller.}
    \label{EPR_N1_Thrust}
\end{figure}

The comparison of thrust response for the $N_1-N_2$ and $\Pi_\mathrm{EPR}-N_1$ strategies is depicted in Fig. \ref{EPR_N1_Thrust}. As can be seen, the EPR-based strategy leads to smaller overshoots and a less oscillatory response in the circled regions. However, the overshoots are still apparent for the AKMPC due to quick changes of $A_\mathrm{n}$, affecting the thrust directly via the $A_\mathrm{n}(p_\mathrm{ex} - p_0)$ term in (\ref{eq_thrust}). The K-FBLC $\Pi_\mathrm{EPR}-N_1$ control exhibits the overall best thrust response. 

\begin{table}[!ht] 
    \caption{The IAE criterion values for the controllers in sea-level conditions.}\label{tab_EPR_SL_IAE}
    \begin{tabular*}{\columnwidth}{@{\extracolsep\fill}l|cc}
    \toprule
    Controller & $\mathrm{IAE}\ N_1$ & $\mathrm{IAE}\ \Pi_\mathrm{EPR}$ \\[2pt]
    Unit & $\mathrm{RPM\,s}$ & (-) \\
    \midrule
    K-FBLC  & 21380 & 3.88 \\[2pt]
    K-FBLC-I  & 21400 & 3.785 \\[2pt]
    AKMPC & 20922 & 3.738 \\
    \bottomrule
    \end{tabular*}
\end{table}

\subsubsection{Varying Flight Conditions}
The same altitude and Mach number profile was utilized, and the step profile was now considered for the EPR. The corresponding $N_1$ was read from the steady-state characteristics.

\begin{figure}
    \centering
    \includegraphics[width=1\linewidth]{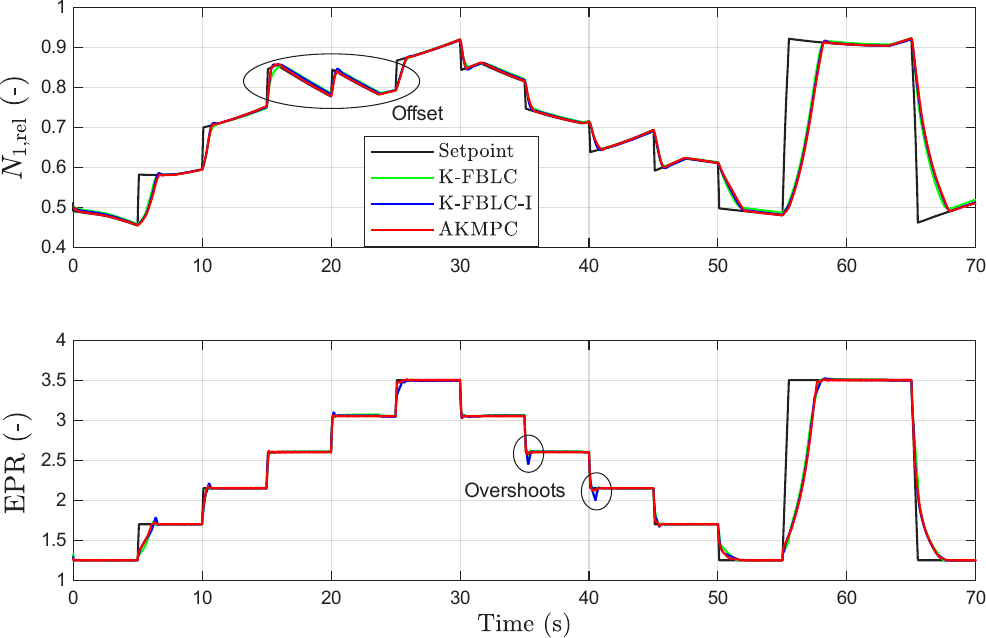}
    \caption{EPR and $N_1$ response in varying flight conditions. Black circles mark the regions with offset and overshoots.}
    \label{EPR_N1_VFC_EPR}
\end{figure}

\begin{figure}[!ht] 
    \centering 
    \begin{subfigure}{0.45\textwidth}
        \centering
        \includegraphics[width=0.9\linewidth]{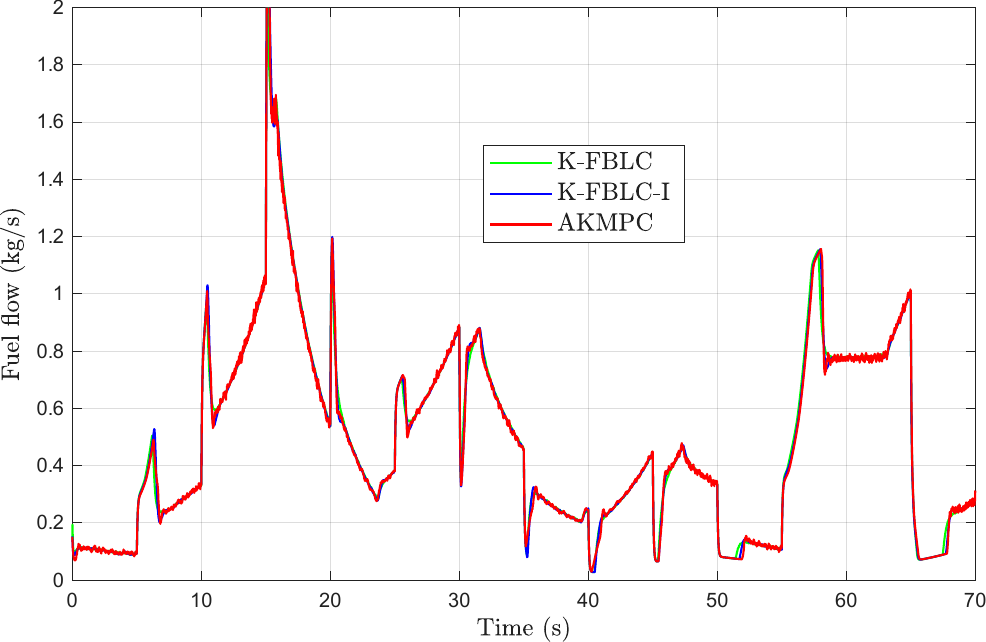}
        \caption{}
        \label{EPR_N1_VFC_Wf}
    \end{subfigure}
    \begin{subfigure}{0.45\textwidth}
        \centering
        \includegraphics[width=0.9\linewidth]{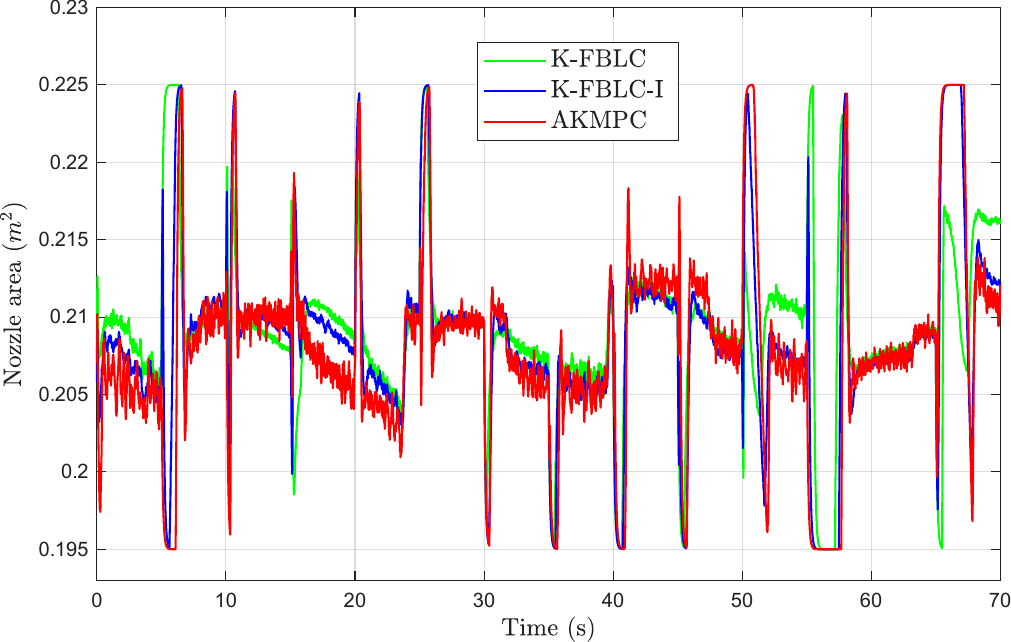}
        \caption{}
        \label{EPR_N1_VFC_An}
    \end{subfigure}
    \caption{(a) Fuel flow and (b) nozzle area comparison for $\Pi_\mathrm{EPR}-N_1$ in varying flight conditions.}
    \label{EPR_N1_VFC_Inputs}
\end{figure}

The results are shown in Fig. \ref{EPR_N1_VFC_EPR} and Table \ref{tab_EPR_VFC_IAE}. The AKMPC adheres better to the setpoint and captures the effects of changing flight conditions thanks to the DO. The K-FBLC also tracks the EPR setpoint relatively accurately with a marginally higher IAE, overshoots, and a steady-state offset. The $N_1$ tracking performance is worse, with an offset in some regions. The offset is successfully mitigated by the integrators for the K-FBLC-I at the cost of slightly increased overshoots (circled in the figure). Analyzing the inputs in Fig. \ref{EPR_N1_VFC_Inputs}, the same trend can be observed as in the sea-level conditions, especially for large steps. 

\begin{figure}[!ht]
    \centering
    \includegraphics[width=0.9\linewidth]{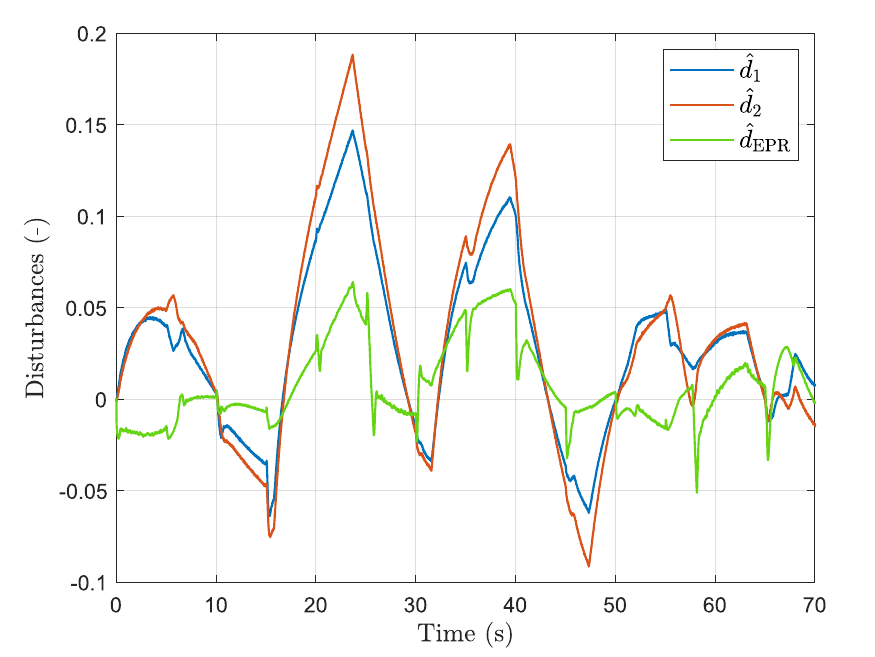}
    \caption{Estimated disturbances for the $\Pi_\mathrm{EPR}-N_1$ case in varying flight conditions.}
    \label{EPR_N1_VFC_Dist}
\end{figure}

The estimated disturbances are shown in Fig. \ref{EPR_N1_VFC_Dist}. Compared to Fig. \ref{N1_N2_VFC_Dist}, the corrections were again smaller, with a maximum of 0.188.

\begin{table}[!ht] 
    \caption{The IAE criterion values for the controllers in varying flight conditions.}\label{tab_EPR_VFC_IAE}
    \begin{tabular*}{\columnwidth}{@{\extracolsep\fill}l|cc}
    \toprule
    Controller & $\mathrm{IAE}\ N_1$ & $\mathrm{IAE}\ \Pi_\mathrm{EPR}$ \\[2pt]
    Unit & $\mathrm{RPM\,s}$ & (-) \\
    \midrule
    K-FBLC  & 25366 & 4.703 \\[2pt]
    K-FBLC-I  & 24500 & 4.55 \\[2pt]
    AKMPC & 23807 & 4.331 \\
    \bottomrule
    \end{tabular*}
\end{table}

\section{Conclusion} \label{sec7}
In this paper, Koopman operator-based approaches were investigated for multivariable control of a turbofan GTE. A data-driven identification framework based on MH-EDMD was employed, with a modified multi-criteria cost function designed to ensure accurate prediction of spool speeds and EPR. The identified time-varying low-order Koopman model was suitable for multiple control strategies, including the considered $N_1-N_2$ and $\Pi_\mathrm{EPR}-N_1$ configurations. A diversity metric based on the Euclidean distance between observables in the parameter space was also included in the objective. Among the tested basis functions, the IQ RBF provided the best prediction performance. Additional output variables, such as thrust, surge margin, or exhaust temperature, can be incorporated into the identification objective provided that reliable training data or validated estimators are available. 

The MH-EDMD was compared with a classical EDMDc approach. This comparison showed the benefit of using state-dependent input dynamics and optimizing the nonlinear observable parameters. While the EDMDc model relies on a fixed dictionary and LTI input dynamics, the proposed approach provides a time-varying linear Koopman representation suitable for prediction and control across the investigated operating range.

Using the identified Koopman model, three control approaches were developed: the AKMPC with a DO, the K-FBLC, and the K-FBLC-I. The AKMPC can be interpreted as a computationally efficient alternative between the offset-free KMPC with a fixed model and online adaptive Koopman identification. The Koopman model is identified offline, its state-dependent matrices are evaluated online at the current operating point, and mismatch is compensated via the DO.

The simulation results showed that the AKMPC provides the most consistent performance across both sea-level and varying flight conditions. Under sea-level conditions, the AKMPC and K-FBLC-based controllers achieved comparable tracking performance in several cases. However, under varying flight conditions, the AKMPC demonstrated improved robustness due to the disturbance observer and predictive constraint handling. The integrators in K-FBLC-I reduced steady-state offsets, but they led to occasional overshoots and retained the structural dependence on the selected output relative degree. For the $\Pi_\mathrm{EPR}$--$N_1$ strategy, both AKMPC and K-FBLC-based controllers were able to track the EPR reference, confirming that the identified Koopman model captures the relevant EPR dynamics. Nevertheless, the K-FBLC requires structural modifications when the controlled outputs or the relative degree change, whereas the AKMPC requires only modifications to the output prediction matrices. The Koopman-based EPR control also leads to improved thrust response.

The computational-time analysis indicated that the online AKMPC update, including Koopman prediction and QP, is computationally tractable in the tested MATLAB implementation. However, the final execution time would depend on the target hardware, operating system, language, and selected QP solver.

An important advantage of the AKMPC formulation is its ability to include GTE output limiters directly as linear inequality constraints. In addition to the input constraints, which were estimated based on the surge and temperature limits, explicit output constraints were also evaluated. The presented results demonstrated the effect of these constraints on the closed-loop response. In particular, quantities such as rotor acceleration rates and turbine inlet temperature can be predicted using the Koopman model and incorporated into the optimizer over the prediction horizon. This highlights a relevant safety-oriented benefit of the Koopman representation: limited nonlinear outputs can be represented in a form suitable for constrained MPC.

The properties of the proposed AKMPC should also be interpreted in the context of the identified Koopman model. The MH-EDMD objective explicitly penalizes prediction error accumulated over the training time series, which supports reliable prediction over the substantially shorter AKMPC prediction horizons. The output DO then compensates output mismatch and flight-condition variations. The present study focuses on practical closed-loop behavior over the investigated operating envelope rather than on a theorem-based stability analysis. The controller solves a constrained finite-horizon QP at each sampling instant. Thus, the imposed input and output constraints and penalties, disturbance compensation, and the simulation results support practical closed-loop stability over the investigated operating envelope.

Regarding the limitations, the output constraints rely on the accuracy of the Koopman predictions. Therefore, constraint tightening should be applied to account for prediction error and uncertainty. Also, direct use of TIT or surge constraints requires reliable estimation of these quantities, either from measurements, laboratory-calibrated estimators, or validated engine models. Furthermore, the DO only accounts for output disturbances. However, state disturbances may be included as well, provided that their effect on the state dynamics can be represented, yielding an extended-state-observer formulation. Finally, the proposed approach keeps the observable functions fixed and updates only the state-dependent matrices online. Consequently, large changes outside the identified operating envelope may require re-identification or extension of the training data.

Future work will focus on extending the framework to additional control variables, such as variable bleed valves or afterburner fuel flow, application to different GTE configurations, e.g., variable-cycle engines, integration of Koopman-based models for direct thrust control in combination with reliable thrust estimation, and conducting experimental validation on small-scale engines.

\bmhead{Acknowledgements}

\bmhead {Author contribution}
The author confirms sole responsibility for the following: study conception and design, simulation and data collection, analysis and interpretation of results, and manuscript preparation.

\bmhead {Funding}
The author declares that this research was supported by the infrastructure of the University of Defence, Brno, Czech Republic, within the framework of DZRO-FVT22-AIROPS “Long Term Organization Development Plan - Conduction of airspace operations”, and by the Student Research Program of the Ministry of Education, Youth and Sports of the Czech Republic.

\bmhead{Data availability}
Data will be made available on reasonable request.

\bmhead{Artificial intelligence}
ChatGPT and Grammarly were utilized to improve the quality of this paper, particularly for editing and language checking. The author declares that he did not use the AI tools for the development of the methods and interpretation of the results.

\section*{Declarations} \label{sec8}

\bmhead {Conflict of interest}
The author has no conflicts of interest and no relevant financial or non-financial interests to disclose.

\bibliography{bibliography}

\end{document}